\documentclass[sigconf]{acmart}

\AtBeginDocument{%
  \providecommand\BibTeX{{%
    \normalfont B\kern-0.5em{\scshape i\kern-0.25em b}\kern-0.8em\TeX}}}

\acmConference[CVMP 2023]{the 20th ACM SIGGRAPH European Conference on
Visual Media Production}{Nov.\ 30--Dec.\ 1}{London, UK}
\acmYear{2023}
\copyrightyear{2023}
\setcopyright{none}

\usepackage{times}
\usepackage{epsfig}
\usepackage{graphicx}
\usepackage{amsmath}

\usepackage{comment}
\usepackage{nicefrac}
\usepackage{booktabs}
\usepackage{multirow}
\usepackage{colortbl}

\usepackage{colortbl}
\usepackage{multicol}
\usepackage{wrapfig}
\usepackage[labelfont=bf]{caption}
\usepackage{tabularx}
\usepackage{utfsym}
\usepackage{soul}
\usepackage{listings}
\usepackage{subcaption}

\definecolor{Yellow}{rgb}{1.0, 1.0, 0.6}
\definecolor{Orange}{rgb}{1.0, 0.8, 0.6}
\definecolor{Red}{rgb}{1.0, 0.6, 0.6}
\definecolor{brightpink}{rgb}{1.0, 0.0, 0.5}
\definecolor{jade}{rgb}{0.0, 0.66, 0.42}
\definecolor{pastelbrown}{rgb}{0.51, 0.41, 0.33}

\definecolor{brightpink}{rgb}{1.0, 0.0, 0.5}
\definecolor{brightgreen}{rgb}{0.4, 1.0, 0.0}
\definecolor{cadmiumyellow}{rgb}{1.0, 0.96, 0.0}
\definecolor{canaryyellow}{rgb}{1.0, 0.94, 0.0}
\definecolor{caribbeangreen}{rgb}{0.0, 0.8, 0.6}
\definecolor{citrine}{rgb}{0.89, 0.82, 0.04}
\definecolor{chromeyellow}{rgb}{1.0, 0.65, 0.0}
\definecolor{classicrose}{rgb}{0.98, 0.8, 0.91}
\definecolor{cherryblossompink}{rgb}{1.0, 0.72, 0.77}
\definecolor{carnationpink}{rgb}{1.0, 0.65, 0.79}
\definecolor{candypink}{rgb}{0.89, 0.44, 0.48}
\definecolor{applegreen}{rgb}{0.55, 0.71, 0.0}
\definecolor{ao}{rgb}{0.0, 0.5, 0.0}
\definecolor{brightgreen}{rgb}{0.4, 1.0, 0.0}
\definecolor{caribbeangreen}{rgb}{0.0, 0.8, 0.6}
\definecolor{caribbeangreen2}{rgb}{0.4, 0.8, 0.1}

\definecolor{lightgreen}{RGB}{197,224,180}
\definecolor{lightblue}{RGB}{222,235,247}
\definecolor{lightpurple}{RGB}{238,229,241}
\definecolor{lightorg}{RGB}{251,229,214}
\definecolor{lightorange}{rgb}{1.0, 0.8, 0.6}

\definecolor{keypoints}{HTML}{1864AB}
\definecolor{initial}{HTML}{ED7FA6}
\definecolor{latent}{HTML}{AA6CB9}
\definecolor{camera}{HTML}{087F5B}
\definecolor{virtual}{HTML}{6276D7}
\definecolor{render}{HTML}{0B7285}
\definecolor{projection}{HTML}{087F5B}
\definecolor{inversion}{HTML}{D893E9}
\definecolor{generator}{HTML}{A184F6}
\definecolor{openpose}{HTML}{64AEED}
\definecolor{modnet}{HTML}{5BC3D2}
\definecolor{expose}{HTML}{ED7FA6}
\definecolor{camcalib}{HTML}{59CDAA}

\lstdefinestyle{mystyle}{
    commentstyle=\color{codegreen},
    keywordstyle=\color{magenta},
    numberstyle=\tiny\color{codegray},
    stringstyle=\color{codepurple},
    basicstyle=\ttfamily\footnotesize,
    breakatwhitespace=false,         
    breaklines=true,                 
    captionpos=b,                    
    keepspaces=true,                 
    numbers=left,                    
    numbersep=5pt,                  
    showspaces=false,                
    showstringspaces=false,
    showtabs=false,                  
    tabsize=2
}

\newcommand{\first}[1]{\textbf{#1}\cellcolor{Red}}
\newcommand{\second}[1]{#1\cellcolor{Orange}}
\newcommand{\third}[1]{#1\cellcolor{Yellow}}
\newcommand{\up}[1]{#1$\uparrow$}
\newcommand{\down}[1]{#1$\downarrow$}

\newcommand{\pose}{\boldsymbol{\theta}}
\newcommand{\shape}{\boldsymbol{\beta}}
\newcommand{\transform}{\mathbf{T}}
\newcommand{\rotation}{\mathbf{R}}
\newcommand{\translation}{\mathbf{t}}

\newcommand{\keypoints}{\mathbf{k}}
\newcommand{\joints}{\mathbf{j}}
\newcommand{\error}{\mathcal{E}}
\newcommand{\vertices}{\mathbf{V}}
\newcommand{\faces}{\mathbf{F}}
\newcommand{\generator}{\mathcal{G}}
\newcommand{\latent}{\mathbf{z}}

\newcommand{\regressor}{\mathbf{R}}

\newcommand{\body}{\mathcal{B}}

\newcommand{\mesh}{(\vertices, \faces)}
\newcommand{\window}{\mathcal{T}}
\newcommand{\slerp}{\mathcal{S}}
\newcommand{\lerp}{\mathcal{L}}
\newcommand{\manifold}{\mathcal{M}}
\newcommand{\projection}{\boldsymbol{\mathcal{\pi}}}

\newcommand{\ALIAS}[2]{BundleMoCap}

\DeclareMathOperator*{\argmin}{\arg\!\min}

\usepackage{adjustbox}%

\usepackage[capitalize]{cleveref}
\AtEndPreamble{%
\RequirePackage{hyperref}
\hypersetup{%
      colorlinks    = true,%  Color text instead of boxes
      linkcolor     = black,% Color of internal links
      citecolor     = black,% Color of citations
      urlcolor      = magenta,%  Color of external urls
      pdfstartview  = Fit,%
      pdfmenubar    = true,%
      pdftoolbar    = true,%
      bookmarksopen = false,%
    }
}

\copyrightyear{2023}
\acmYear{2023}
\setcopyright{acmlicensed}\acmConference[CVMP '23]{European Conference on
Visual Media Production}{November 30-December 1, 2023}{London, United
Kingdom}
\acmBooktitle{European Conference on Visual Media Production (CVMP '23),
November 30-December 1, 2023, London, United Kingdom}
\acmPrice{15.00}
\acmDOI{10.1145/3626495.3626511}
\acmISBN{979-8-4007-0426-0/23/11}

\renewcommand\footnotetextcopyrightpermission[1]{} %
\begin{document}

\title{BundleMoCap: Efficient, Robust and Smooth Motion Capture from Sparse Multiview Videos}

\author{Georgios Albanis}
\email{giorgos@moverse.ai}
\affiliation{
\institution{Department of Informatics \& Telecommunications,\\ University of Thessaly}
\city{Lamia}
\country{Greece}
}
\affiliation{
\institution{Moverse}
\city{Thessaloniki}
\country{Greece}
}
\authornote{This is the authors' version of the work. It is posted here for your personal use. The definitive version was published in CVMP '23, https://dl.acm.org/doi/10.1145/3626495.3626511}

\author{Nikolaos Zioulis}
\email{nick@moverse.ai}
\affiliation{
\institution{Moverse}
\city{Thessaloniki}
\country{Greece}
}
\author{Kostas Kolomvatsos}
\email{kostasks@uth.gr}
\affiliation{
\institution{Department of Informatics \& Telecommunications,\\ University of Thessaly}
\city{Lamia}
\country{Greece}
}

\renewcommand{\shortauthors}{G. Albanis et al.}

\begin{abstract}
   Capturing smooth motions from videos using markerless techniques typically involves complex processes such as temporal constraints, multiple stages with data-driven regression and optimization, and bundle solving over temporal windows.
   These processes can be inefficient and require tuning multiple objectives across stages.
   In contrast, BundleMoCap introduces a novel and efficient approach to this problem. It solves the motion capture task in a single stage, eliminating the need for temporal smoothness objectives while still delivering smooth motions. 
   BundleMoCap outperforms the state-of-the-art without increasing complexity.
   The key concept behind BundleMoCap is manifold interpolation between latent keyframes. 
   By relying on a local manifold smoothness assumption, we can efficiently solve a bundle of frames using a single code. 
   Additionally, the method can be implemented as a sliding window optimization and requires only the first frame to be properly initialized, reducing the overall computational burden.
   BundleMoCap's strength lies in its ability to achieve high-quality motion capture results with simplicity and efficiency.
   For more results visit our project's page \href{https://moverseai.github.io/bundle/}{https://moverseai.github.io/bundle/}%
   
\end{abstract}




\begin{teaserfigure}
  \centering
    \includegraphics[width=\textwidth]{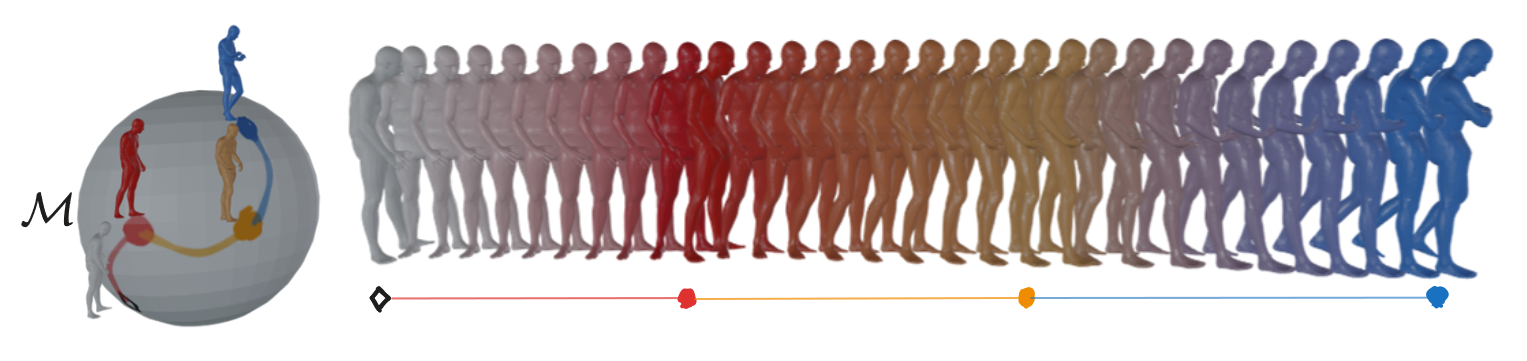}
    \captionof{figure}{
        We introduce a bundle solving approach for motion capture that efficiently produces smooth motion outputs despite fitting to noisy observations, and is robust to outliers.
        Instead of solving for all frames' parameters, BundleMoCap effectively solves for the next latent keyframe and relies on manifold interpolation for reconstructing and constraining a temporal segment.
        Essentially, a single latent code (indicated by the marks on the bottom line on the right) is fit for each bundle of frames that are reconstructed by traversing a pose manifold $\manifold$ from the previously solved latent code (left).
        The entire bundle (color-coded from one keyframe to the next on the right) is solved using constraints for all reconstructed frames.
        BundleMoCap captures smooth motions even when only constrained from a very sparse set of viewpoints that suffer from higher outlier predictions across time.
        More importantly, this requires no additional hyper-parameter tuning as BundleMoCap does not use any temporal smoothness objective.
    }
    \label{fig:teaser}  
\end{teaserfigure}

\maketitle

\section{Introduction}

Human motion capture (MoCap) is a long-standing goal and highly researched topic, with its popularity mainly stemming from the multitude of applications that it can enable in domains such as gaming, content creation, sports, and surveillance.
The complexity of human motion necessitates the use of costly equipment and complex processes but during the last years, the emergence of modern machine learning has created new opportunities to reduce complexity and costs.
Monocular and/or sparse multi-view markerless capture has seen tremendous growth as summarized in a recent state-of-the-art report \cite{tian2023recovering}.

However, most works focus on single image captures or simply extend image-based captures to a video stream of images, producing unrealistic motion streams with a distinct lack of temporal coherence.
While simple localization jitter may be reduced using traditional filtering, occlusion creates challenging deviation patterns that cannot simply be addressed via filtering.
Various solutions \cite{wei2022capturing,zeng2022smoothnet,tang20233d,shen2023global} have been proposed for modelling long-range motion dependencies to overcome this.

Still, these are purely data-driven solutions for the highly ill-posed monocular case.
Sparse multi-view captures are harder to model with end-to-end machine learning models due to the lack of data and the complexities stemming from the limited expressivity and scalability of neural networks when applied to multi-view data.
This has, in turn, led to the predominant use of optimization-based approaches for multi-view MoCap and the integration of motion smoothness objectives in addition to solving entire temporal windows \cite{huang2017towards,huang2021dynamic,arnab2019exploiting,ye2023decoupling}.

Another challenge arises from the data constraints used to solve these optimization problems, typically 2D keypoint estimations from an image-based pose estimation model \cite{openpose}.
These are highly jittery estimates and also suffer from missing or inverted estimates \cite{ruggero2017benchmarking}.
Jittery 2D keypoints accentuate jitter in the captured motion, and missing or inverted estimates render the sparse multi-view setting very sensitive to outliers.
The solution to this in prior work \cite{huang2017towards,ye2023decoupling} has been the use of multi-staged body fits, where each stage progressively refines the solution.
Initial stages typically solve on a per-frame basis, with subsequent stages adding temporal constraints and sometimes solving a bundle of frames simultaneously.
Other solutions involve solving for the entire video simultaneously after extracting initial estimates from a single-shot regressor model \cite{arnab2019exploiting}, or relying on novel motion priors and solving motion segments over multiple stages \cite{huang2021dynamic,rempe2021humor}.

In this work, we present a novel solution for solving sparse multi-view markerless MoCap that addresses all aforementioned challenges.
Instead of solving for a group of frames we solve for a single latent code that reconstructs a bundle of frames via manifold interpolation.
The reconstructed bundle is then constrained and solved across a temporal window.
This approach has several advantages, it is more robust to the prevalent outlier cases that manifest in sparse multi-view settings, it is far more efficient as it solves a video in a single stage, and produces smooth motion with \textbf{\underline{no}} motion smoothness objectives.

\section{Related Work}

\subsection{Markerless Multi-view MoCap} 
In addition to digitizing human motion, capturing the pose and/or motion of humans is critical for dataset creation \cite{cheng2022generalizable,zhang2023neuraldome,huang2022intercap}, training data collection \cite{bhatnagar2020loopreg}, bootstrapping dynamic radiance fields \cite{zhao2022humannerf}, and has even found use in physicalizing human motion \cite{zhang2018mosculp}.
The standard approach is to rely on 2D keypoint detections \cite{openpose} on calibrated images/videos and fit an articulated skeleton to these observations \cite{huang2017towards}.
While there are approaches that only estimate 3D joints \cite{iskakov2019learnable,ye2022faster,bartol2022generalizable}, these do not provide a consistent structure, suffering from varying bone lengths, and neither offer articulation parameters.
Fitting to 2D keypoint detections is challenging even in the multi-view case, especially when the sensors are sparse, as 2D detectors suffer from high amounts of jitter, inversion and missing detections \cite{ruggero2017benchmarking} due to occlusions.
While direct \cite{jiang2022multi} or progressive \cite{gong2023progressive} data-driven regression approaches exist, extra progress is required to improve their performance and generalization.

\subsection{Temporal Constraints}
Capturing motion on a per-frame basis suffers from noisy estimates or fitting to noisy observations.
While temporal filtering is one option to resolve the temporal jitter and inconsistency of per-frame captures \cite{ingwersen2023sportspose}, it is very challenging to design filters to handle occlusions and outliers.
This led to the integration of temporal constraints and priors when solving for the pose and motion.
One of the earlier attempts \cite{kanazawa2019learning} used a bundle of images as input to regressors predicting a coherent bundle of outputs, and specifically the parameters of the SMPL body model.
Follow-up approaches \cite{li2021learning} relied on recurrent architectures which are a better architectural fit to the problem at hand, as each prediction directly depends on the previous frame estimates.
Regarding temporal constraints within optimization loops, the most common objective is a joint smoothness one \cite{arnab2019exploiting,peng2018sfv,zhang2018mosculp,ye2023decoupling} that essentially enforces a position constancy.
A DCT smoothness term has also been used when fitting the SMPL body to multiple images \cite{huang2017towards}.
Another variant involves higher order smoothness terms like velocity constancy \cite{zanfir2018monocular,mosh,amass}, either on joint/marker positions or on joints' angles.
With the advent of data-driven priors, lots of works have added higher level smoothness constraints to supplement the joint position one.
These include different representations, like the feature space of a velocity autoencoder \cite{zhang2021learning,huang2022intercap}, the latent space of poses \cite{huang2021dynamic}, motion \cite{saini2023smartmocap}, or the latent space and transition state of an autoregressive motion prior \cite{jin2023robust,ye2023decoupling}. 

A common theme to all prior approaches is that motion smoothness is achieved by the addition of one or more additional objectives to the optimization problem.
This incurs the cost of additional parameters that need tuning, a case that is especially complex when considering the high number of objectives that are involved when fitting a parametric body model to images.
Instead, our approach achieves motion smoothness implicitly by smoothly interpolation a pose manifold, requiring no additional hyper-parameters.

\subsection{Bundle Solving}
Adding such temporal constraints is more effective when solving for a bundle of frames simultaneously.
The DCT prior used in \cite{huang2017towards} was added the second stage of optimization when a group of $30$ frames was solved simultaneously.
Similarly, joint smoothness was imposed on a group of frames solved simultaneously in MoSculp \cite{zhang2018mosculp} as well as other works \cite{peng2018sfv,saini2023smartmocap,ye2023decoupling}.
Bundle solving with joint smoothness is also possible in the latent space of regressor models \cite{peng2018sfv}, albeit costly given its excessive parameter count.
Apart from solving a small bundle, it has also been shown that it is possible to solve entire videos simultaneously \cite{arnab2019exploiting}, exploiting the entire temporal context, even with very complex objective functions \cite{huang2022intercap} involving human-object interactions.
Still this requires good initialization, which can be achieved using a regressor model \cite{arnab2019exploiting} or a per-frame body fitting stage \cite{huang2022intercap}.
This type of staged optimization is standard in the ill-posed monocular case \cite{bogo2016keep,pavlakos2019expressive}, but also finds widespread use in multiview settings \cite{huang2017towards,saini2023smartmocap}.
When considering videos, multiple stages are used to initialize estimates for each separate frame, and then solve bundles using smoothness constraints \cite{zhang2021learning,ye2023decoupling,saini2023smartmocap,arnab2019exploiting,peng2018sfv}, be it either via direct regression models or solving an optimization problem for each frame.

Nonetheless, this is rather costly, as multiple passes are required over the videos. 
In addition, the temporal nature of the data are only considered at the constraints level, but not at the parameters level, with each frame's state being solved separately in the bundle.
Our approach, BundleMoCap, only solves for a single keyframe for each bundle, and only requires a single -- first -- frame initialization, essentially solving for smooth motion robustly in a single stage.

\subsection{Latent Parameters}
Earlier body fitting works solved for the pose articulation parameters directly and regularized the solution to plausible poses using Gaussian mixture priors \cite{bogo2016keep,huang2017towards}.
Modern representation learning offered a more compact latent space to solve instead of the rotation parameter space and simultaneously allowed for penalizing implausible poses directly on the solved parameters \cite{pavlakos2019expressive}.
As a result, while earlier bundle solving MoCap works solved for the articulation \cite{arnab2019exploiting,huang2017towards,zanfir2018monocular}, solving for latent representations is now established \cite{pavlakos2019expressive,ye2023decoupling,saini2023smartmocap}.
Even though initial attempts used the latent space of regressors that is not compact \cite{peng2018sfv}, or pure autoencoder embeddings \cite{zhang2021learning}, variational autoencoders (VAE) \cite{kingma2015iclr} have proven to be more effective as a prior over their latent space enforces the solution to be directed towards more plausible poses.
These have been used to model poses, as in the case of VPoser \cite{pavlakos2019expressive}, as well as motion \cite{saini2023smartmocap,huang2021dynamic}, with the latter typically employing higher dimensional latent spaces. 
Another category of solved latent spaces are autoregressive models, represented by HuMoR \cite{jin2023robust}, which has been used when in video-based MoCap \cite{ye2023decoupling} during the third solving stage to offer motion smoothness. 
Still, a VPoser VAE was used in the second stage for pose plausibility. 
On the other hand, GAN-based methods\cite{davydov2022adversarial} discriminate the generated poses from the real ones, essentially learning adversarial priors, while Pose-NDF \cite{tiwari2022pose} employs a neural implicit representation for modelling the plausible pose manifolds with a neural implicit function. 

In BundleMoCap we show that a compact pose prior can be used to solve for smooth motion using a single latent code, instead of relying on motion or autoregressive priors.
This carries the advantage of being less reliant on data offering a high variety of motion transitions, and more reliant on learning a high quality smooth pose manifold, reducing data pressure.

\section{BundleMoCap}

\subsection{Preliminaries}

To capture human motion, we use a parametric human body model $\body$ that acts as a function over a group of parameters to generate a human body geometry $\mesh = \body(\shape, \pose, \transform)$. A triangular mesh surface $\mesh$ is defined by the vertices $\vertices \in \mathbb{R}^{V \times 3}$ and faces $\faces \in \mathbb{N}^{F \times 3}$. It is reconstructed by $\body$ using $S$ blendshape coefficients $\shape \in \mathbb{R}^S$, articulated by $P$ pose parameters $\pose \in \mathbb{SO}(3)^P$, and globally positioned by the transform $\transform = \left[ \begin{smallmatrix} \rotation & \translation\\ \mathbf{0}&1 \end{smallmatrix} \right] \in \mathbb{SE}(3)$.
Using linear operations expressed as a matrix $\regressor \in \mathbb{R}^{J \times V}$ it is possible to extract $J$ different body joints $\joints \in \mathbb{R}^{J \times 3}$ via matrix multiplication $\joints = \regressor \times \vertices$.
The joints $\joints$ are projected to an image domain $\Omega := \mathbb{R}^{W \times H}$ as keypoints $\keypoints = \projection(\joints)$ using a projection function $\projection$ parameterized by the intrinsic properties of $\Omega$.

\begin{figure*}[!htbp]
\includegraphics[width=\textwidth]{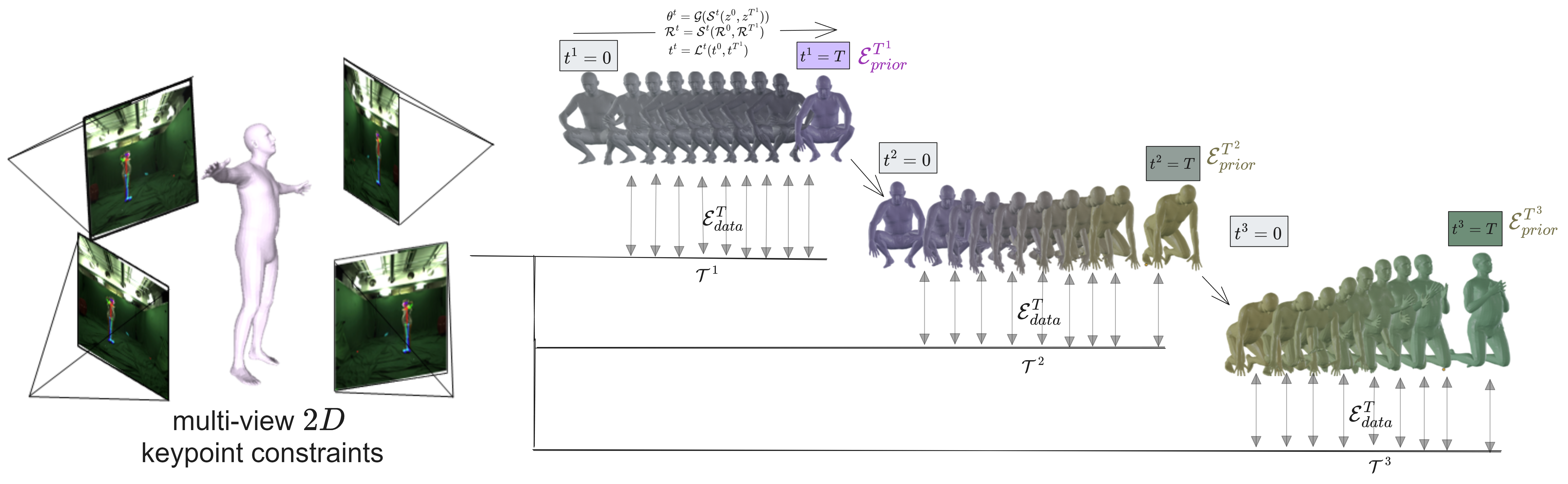}
\caption{
BundleMoCap fits an articulated template mesh to 2D keypoint observations from a sparse set of multi-view videos.
Instead of iteratively optimizing pose parameters for each frame, we focus on optimizing the latent code $\latent^t$ corresponding to the pose parameters $\theta^t = \generator(\latent^t)$ for a single keyframe ($t^i=T$). 
This relies on the reconstruction of the poses, root orientation and translation via interpolation, generating the intermediate frames (visually indicated by the blending between the start and end keyframes).
A sliding window optimization implementation is used where only the first frame is fit in a standalone manner.
Then, the $i\text{th}$ temporal window $\window^i$ that is solved as a bundle, optimizes only the next latent keyframe ($t^i=T$), while reconstructing the frames between the previously optimized keyframe ($t^i=0$) and the next.
All reconstructed frames are constrained by the respective multi-view keypoint constraints via $\error^\window_{data}$ while the latent keyframe is regularized by $\error^\window_{prior}$.
BundleMoCap requires just a single stage to achieve comparable results with the state-of-the-art without pose initialisation for each frame and delivers smooth motions efficiently without using any motion smoothness objective.
}
\label{fig:approach}
\end{figure*}

\subsection{Latent Keyframe Bundle Solving}

To estimate the human pose and shape at a single time instance $t$ a minimization problem is formulated using an objective comprising weighted data $\mathcal{E}_{data}$ and prior $\mathcal{E}_{prior}$ terms \cite{bogo2016keep,pavlakos2019expressive}:
\begin{equation}
\label{eq:optimization}    
    \argmin_{\latent^t,\shape^t,\transform^t} \mathcal{E}^t_{data} + \mathcal{E}^t_{prior},
\end{equation}
with a notable difference being the optimization over a latent code $\latent \in \mathbb{R}^L$ instead of the pose parameters $\pose$.
While earlier works \cite{bogo2016keep} relied on Gaussian mixture pose priors using the joints' angles $\pose$ as input, recent works are now using data-driven priors \cite{pavlakos2019expressive} to instead optimize lower dimensional latent codes.  
These reconstruct the pose parameters $\pose = \generator(\latent)$, with the generator function $\generator$ being a fixed, pre-trained neural network.
Apart from the benefit of optimizing a lower dimensional parameter space, it is also possible to impose a prior term on the latent code itself \cite{pavlakos2019expressive,davydov2022adversarial,tiwari2022pose}.
This is important as it helps prevent degenerate solutions and provides additional constraints to alleviate the ill-posedness of the problem.
Another use of such priors is to initially improve the convexity of the problem when weighted higher, and stage the solve by progressively lowering the prior weights as the solution is refined.
This type of annealing is typical and used in both challenging monocular \cite{bogo2016keep,pavlakos2019expressive} or multi-view cases \cite{huang2017towards}, and even when fitting body models to marker sets \cite{mosh,amass}.
Still, this multi-staged solving process incurs higher run-times that are costly as each time instance $t$ is separately solved and is usually improved by fixing the shape $\shape$ using a single estimate across the entire sequence.

Our proposition is to solve a temporal window $\window := [0, \dots, T]$ simultaneously using two latent keyframes at time instances $0$ and $T$:
\begin{equation}
\label{eq:bundle_optimization}    
    \argmin_{\latent^0, \latent^T, \transform^0, \transform^T} \mathcal{E}^\window_{data} + \mathcal{E}^{0, T}_{prior},
\end{equation}
using an initial and fixed shape estimate $\shape$, with the prior term imposed only on the keyframes, and the data term defined over the entire window $\window$:
\begin{equation}
\label{eq:data_term_window}    
    \mathcal{E}^\window_{data} = \sum\limits_{t=1}^{T} \mathcal{E}^t_{data}.    
\end{equation}

where:
\begin{equation}
 \label{eq:reprojection}
\mathcal{E}^t_{data} = \lambda_R\sum_{c}^{C}\sum_{i}^{J}{ w_i \rho(\mathbf{k}^{t}_{i} - \mathbf{k}^{t}_{det, i}) }
\end{equation}

Here, $\rho$ is the Geman-McClure penalty function which we favour over a traditional L2 as it can deal better with noisy estimates.

To solve over the entire window and fit our solution to all available data constraints, we reconstruct the intermediate frames via interpolation:

\begin{equation}
\label{eq:reconstruction2}
  \pose^t = \generator(\mathcal{S}^t(\latent^0, \latent^T)), \! \left[ \!
  \begin{smallmatrix} \rotation^t & \translation^t \\ \mathbf{0} & 1 \end{smallmatrix} \! \right] \!\! = \!\! \left[ \! \begin{smallmatrix} \mathcal{S}^t(\rotation^0, \rotation^T) & \mathcal{L}^t(\translation^0, \translation^T)\\ \mathbf{0} & 1 \end{smallmatrix} \! \right],
\end{equation}
with $\slerp^t$ and $\lerp^t$ being spherical and linear interpolation functions that map $t$ to a closed unit interval ($[0, 1]$) within $\window$ to blend the starting $t = 0$ and end $t = T$ points, as depicted below:

\begin{equation}
        \mathcal{S}^t = \frac{\sin{\big(1 - \frac{t}{T}\big)\theta}}{\sin{\theta}} \latent^0 + \frac{\sin{\frac{t}{T}\theta}}{\sin{\theta}} \latent^T
    \label{eq:slerp}
\end{equation}
with $\theta$ representing the angular distance between the two latent codes $\latent^0$  and $\latent^T$ \cite{davydov2022adversarial}.

Specifically for the pose $\pose$, this process crucially relies on a well-trained generator $\generator$ that captures an expressive manifold $\manifold$ that can be smoothly interpolated to reconstruct correspondingly smooth pose space transitions.

\section{Results}
\subsection{Implementation Details.}

We use SMPL-X \cite{pavlakos2019expressive} as the parametric human body model $\body$. 
Even though we only optimize for the body joints and ignore the expressive parts, \textit{i.e.}~the hands and face, we use SMPL-X as its corresponding pose prior, VPoser \cite{pavlakos2019expressive}, offers a well-trained decoder/generator $\generator$ over the manifold $\manifold$ of plausible poses.
To obtain the 2D keypoint constraints, $\keypoints_{det}$ and the confidence score $w_i$ for the $i^{th}$ joint, we use OpenPose \cite{openpose}, an established 2D keypoint estimator model that predicts human joint positions in the 2D image space.
We solve the optimization problem using the limited memory BFGS (L-BFGS) optimizer \cite{wright1999numerical}, with a strong Wolfe line search strategy to determine suitable step sizes during optimization.
Optimization iterations are performed using a budget of $30$ iterations, striking a balance between convergence speed and computation resources.
All experiments are implemented with a custom PyTorch-based \cite{paszke2017automatic} framework \cite{moai} using the same PyTorch version, specifically \textit{1.12}. 
We set the weight for the data term and prior term to $1.0$ and $10.76$ respectively and optimize only for a single stage, using a temporal window of length $|\window| = 10$.

\subsection{Sliding Window Optimization}
Our formulation allows for an efficient sliding window optimization implementation. 
Instead of optimizing Eq.~\eqref{eq:bundle_optimization} over the two latent keyframes, we can only optimize for the next keyframe, \textit{i.e.}~$(\latent^T, \transform^T)$, while keeping the first keyframe, \textit{i.e.}~$(\latent^0, \transform^0)$, fixed.
We initialize the fitting process with a single-frame fit to acquire the first frame ($t=0$) solution for the first temporal window $\window^1$ and then solve for the first window's next latent keyframe $t=T^1$.
The optimization process then slides to the next temporal window $\window^2$ using the latent keyframe $t=T^1$ as the first keyframe of the second temporal window $\window^2$, keeping it fixed, while solving for the next latent keyframe $t=T^2$.
This process repeats until all temporal segments are solved, effectively splitting an $F$ frames sequence into $N = F / T$ temporal windows $\window^i, i \in [0, N]$ of length $|\window^i|=T+1$ represented as $N+1$ latent keyframes $(\latent^i, \transform^i)$.
For the first frame we solve Eq.~\eqref{eq:optimization} with the same process and weights using $2$ stages of optimization.
This initial single frame fit is initialized using an HMR \cite{kanazawa2018end} prediction, and also initializes the shape parameters $\shape$.
The latter are kept constant for the entire sequence.
This process is consistent for all methods that we will compare our method to, with any method relying on regressor estimates using HMR, and the same shape parameters used for all methods.

\subsection{Experimental Setup}
\subsubsection{Datasets}
We use two standard multi-view datasets for assessing the performance of our approach.
They offer a high level of poses variance, ranging from simple performances to challenging motions.

Human3.6M \cite{h36m_pami} is a large-scale dataset for 3D human pose estimation, including $3.6M$ video frames from four synchronized cameras and 3D body joint annotations acquired from an optical MoCap system marker-based. 
It includes $11$ human subjects (five females and six males), and according to previous works [\cite{arnab2019exploiting}], S1, S5, S6, S7, and S8 are used for training and S9, and S11 for testing.

MPI-INF-3DHP \cite{mono-3dhp2017} is a dataset for 3D human pose estimation and is obtained through the multi-camera marker-less MoCap system. 
Since its test data includes single-view images, only train data composed of multi-view (\textit{i.e.} $14$) images are used in our experiments. 
Following prior works, one subject is used for testing (S8) out of the total eight captured subjects. 
Likewise, the views numbered $0$, $2$, $7$, and $8$ are used for our experiments.

\subsubsection{Metrics}
Performance is assessed over a range of error and accuracy metrics.
The typically reported MPJPE evaluates joint position estimation error, accompanied by a joint level RMSE that accentuates higher errors.
Since we capture the full articulation of the subject using a template mesh, we also report the angular error (MAE) of the kinematic chain's rotations.
Finally, the accuracy of the estimates is reported with distance thresholded success (PCK) using two different thresholds set at $3$ and $7$cm.

\subsubsection{Methods}
As a baseline method we use standard multi-view fitting that considers each frame separately.
We adapt a multi-staged fitting approach \cite{huang2017towards} ($MuVS$), using a data-driven prior, VPoser \cite{pavlakos2019expressive}, instead of optimizing directly over the angular domain and relying on a Gaussian Mixture Model (GMM) for pose plausibility.
In addition, we initialize the solution with a data-driven regressor estimate from HMR \cite{kanazawa2018end}.

Then a variety of bundle solving methods are compared with our approach.
We start from a multi-staged multi-view MoCap method \cite{huang2017towards} (DCT) that solves for a bundle of frames simultaneously at the last stage using a low frequency basis prior for enforcing smoothly varying joint positions. 
Another approach exploits the entire temporal context \cite{arnab2019exploiting} (ETC) and solves for all frames' parameters simultaneously using a joint smoothness objective, starting from HMR predictions.
ETC is adapted from the monocular case to our experimental multi-view setting.
Then, we use the bundle solving approach of DMMR \cite{huang2021dynamic} using a temporal VPoser motion prior.
While DMMR solves for the camera poses as well, in our experiments the camera poses are known and fixed, ensuring a proper and isolated bundle solving comparison.
DMMR achieves motion smoothness by constraining the nearby optimized latent codes to be close.
Similarly, SLAHMR \cite{ye2023decoupling} also solves for the camera's motion, but this part is skipped in our experiments, instead using the known extrinsic camera calibration parameters, similar to DMMR.
SLAHMR uses a multi-stage fitting approach relying on both VPoser, as well as HuMoR, an autoregressive model for solving over a temporal window and enforcing smooth motion with a joint smoothness objective.
Its constraints are straightforwardly adapted from single view to multi-view for our experiments.

\subsection{Discussion}

\begin{figure*}[!htp]

\begin{subfigure}{0.18\textwidth}
    \includegraphics[width=\textwidth]{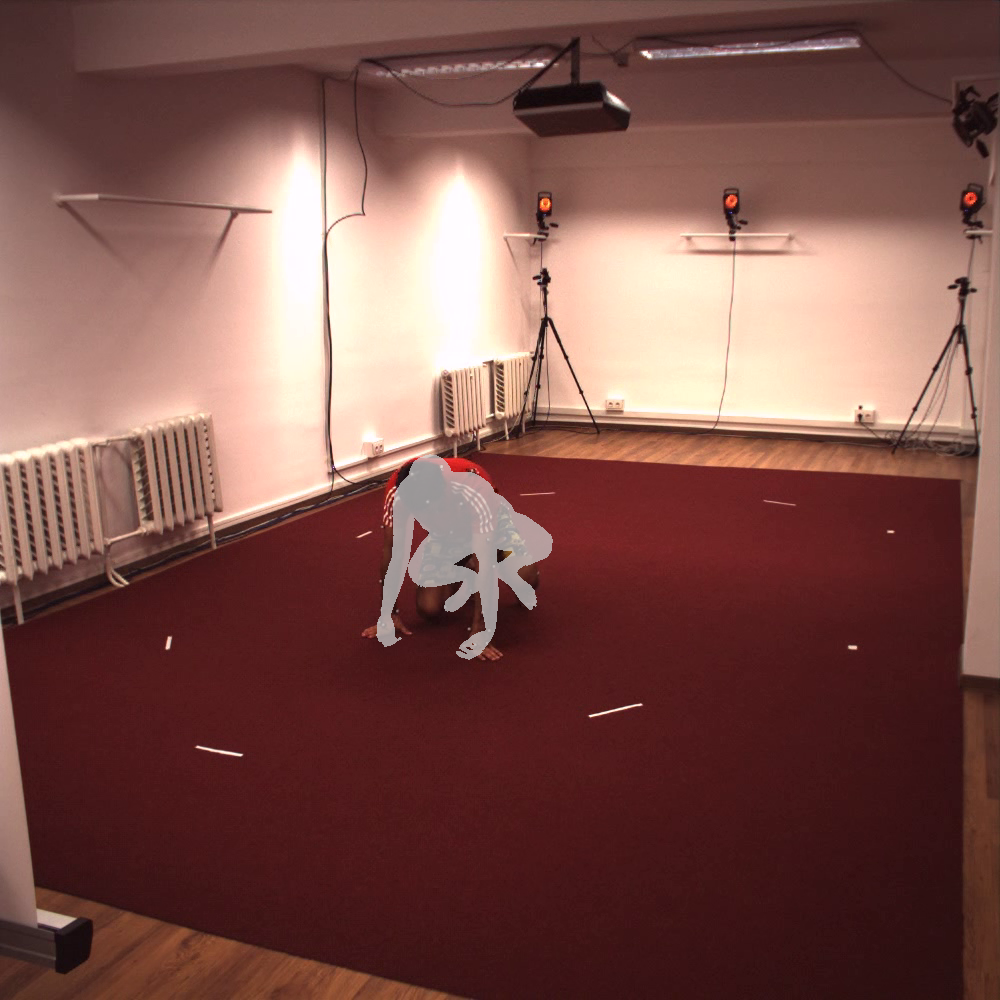}
\end{subfigure}\hfill
\begin{subfigure}{0.18\textwidth}
    \includegraphics[width=\textwidth]{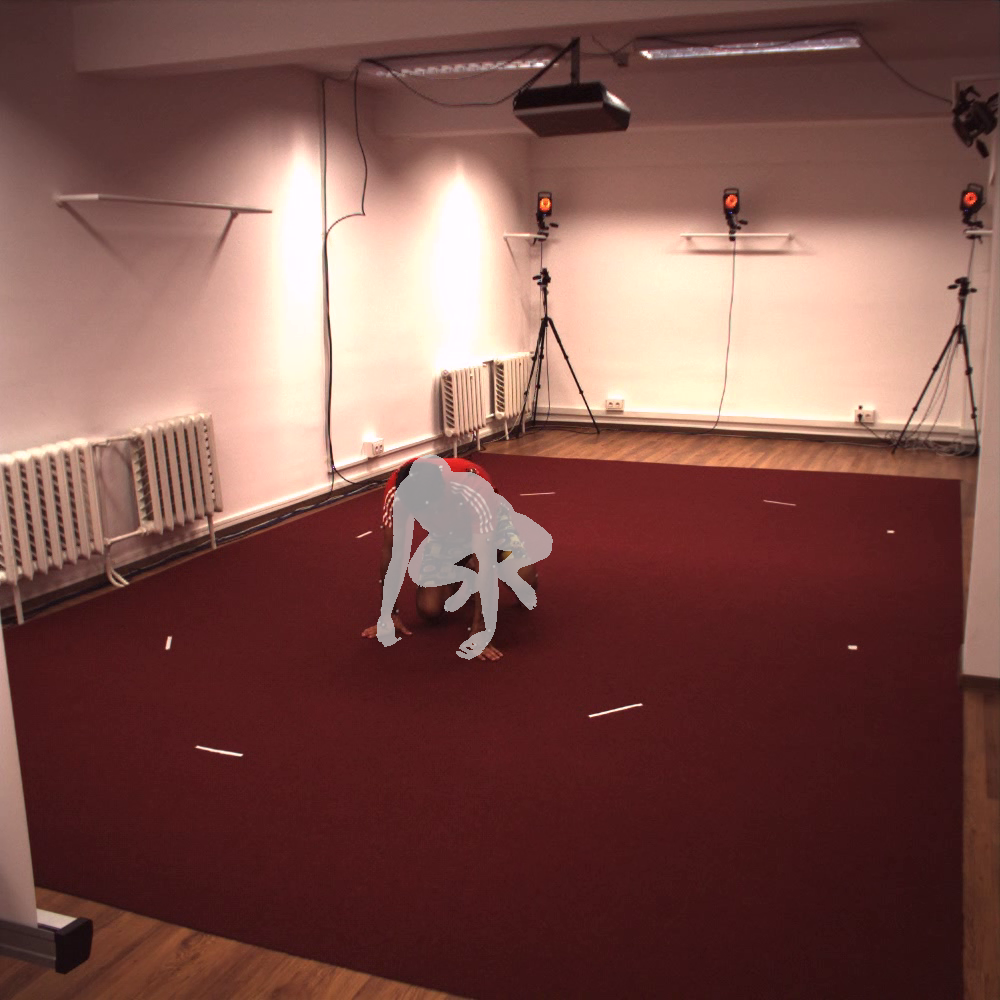}
\end{subfigure}\hfill
\begin{subfigure}{0.18\textwidth}
    \includegraphics[width=\textwidth]{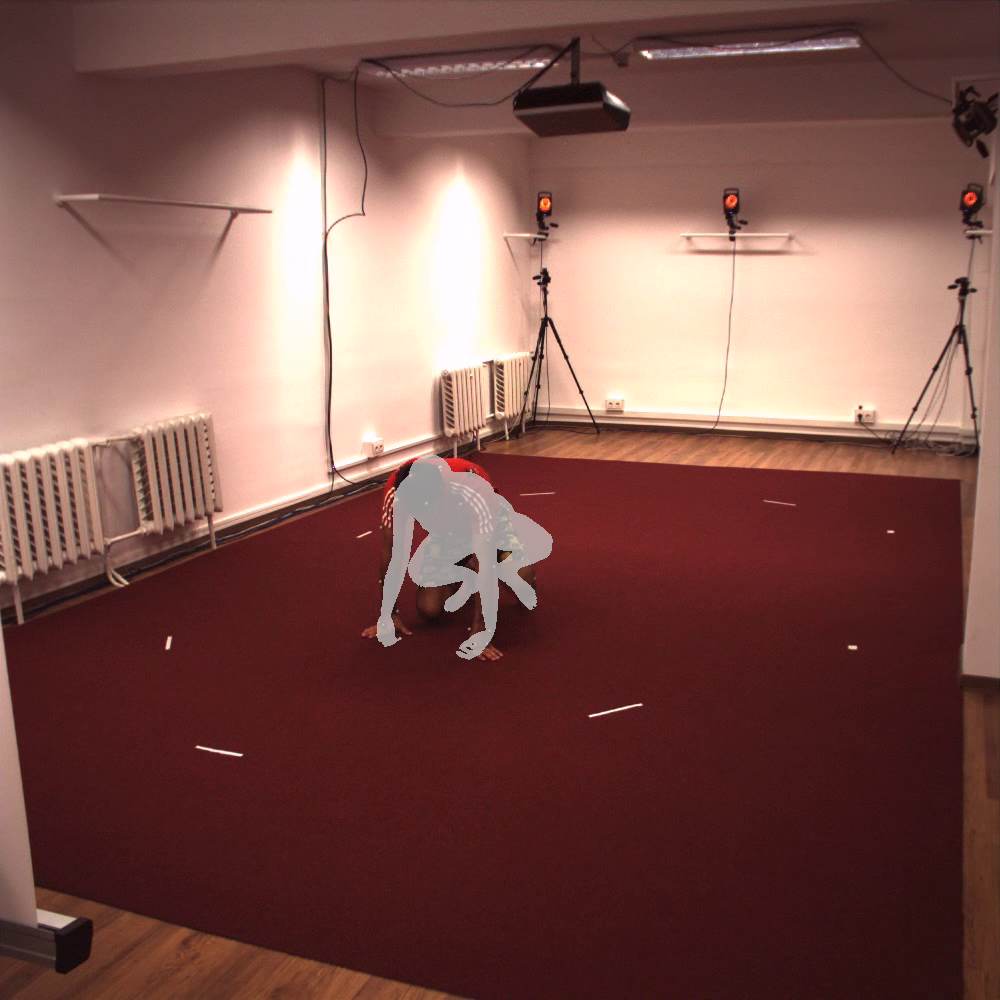}
\end{subfigure}\hfill
\begin{subfigure}{0.18\textwidth}
    \includegraphics[width=\textwidth]{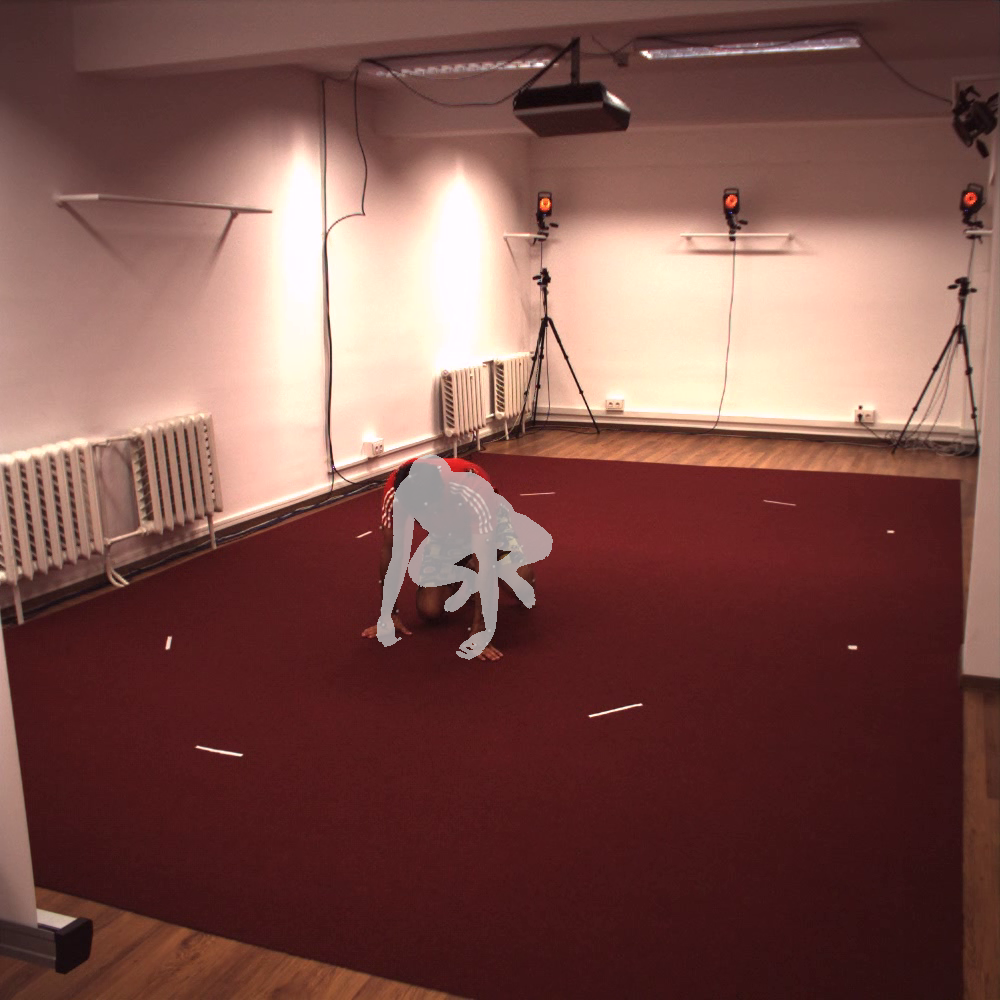}
\end{subfigure}\hfill
\begin{subfigure}{0.18\textwidth}
    \includegraphics[width=\textwidth]{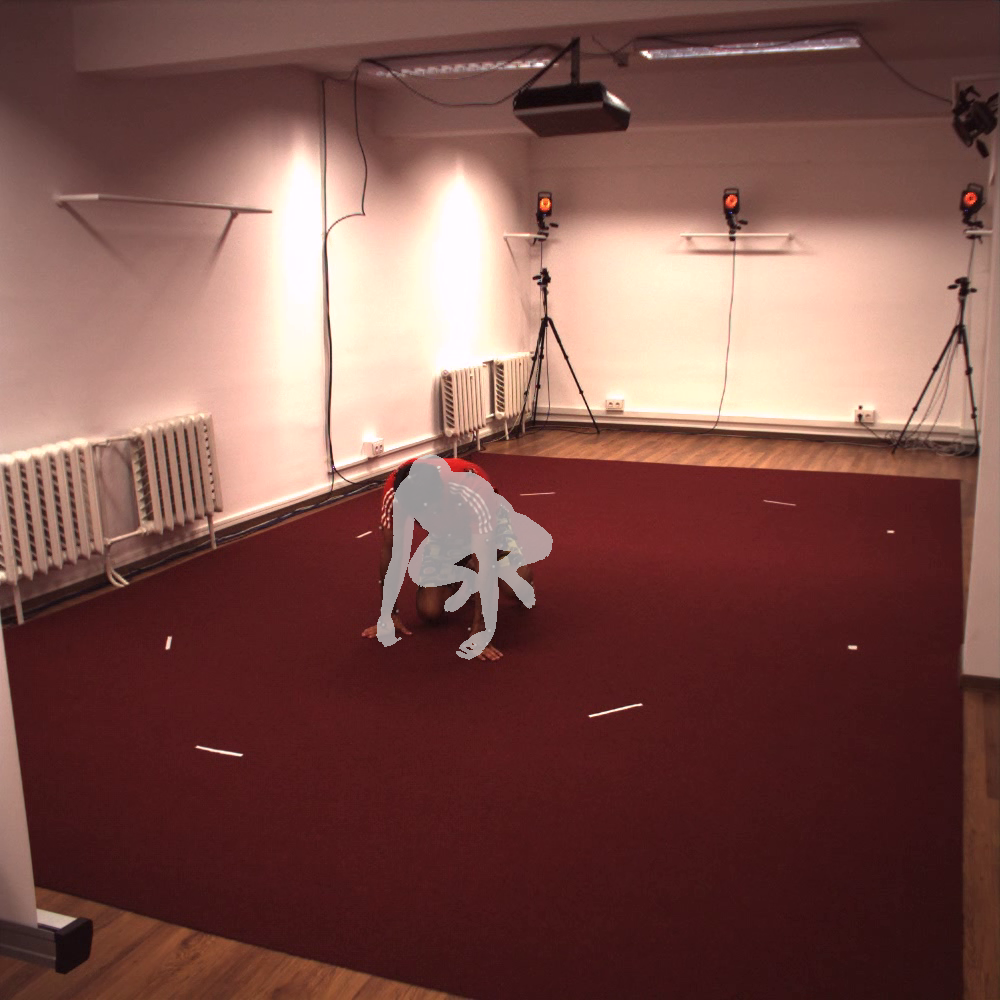}
\end{subfigure}

\begin{subfigure}{0.18\textwidth}
    \vspace{5pt}
    \includegraphics[width=\textwidth]{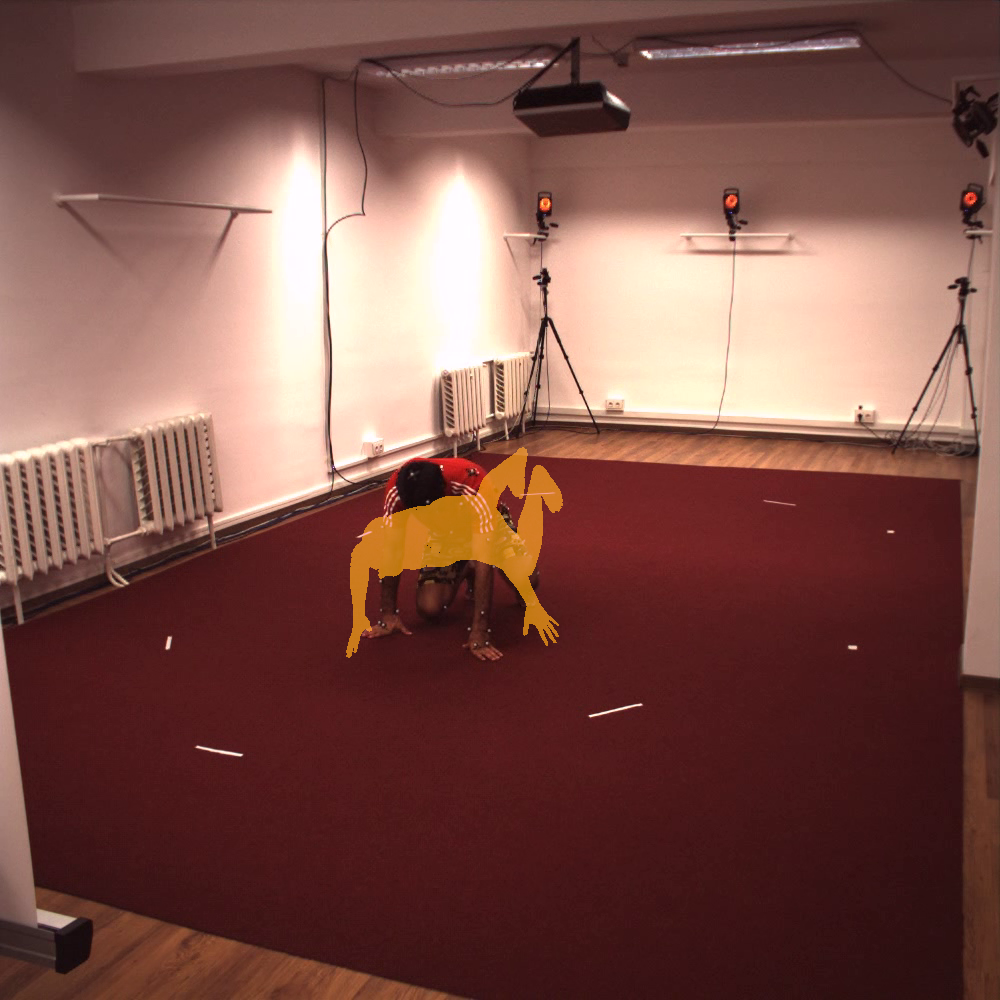}
\end{subfigure}\hfill
\begin{subfigure}{0.18\textwidth}
    \vspace{5pt}
    \includegraphics[width=\textwidth]{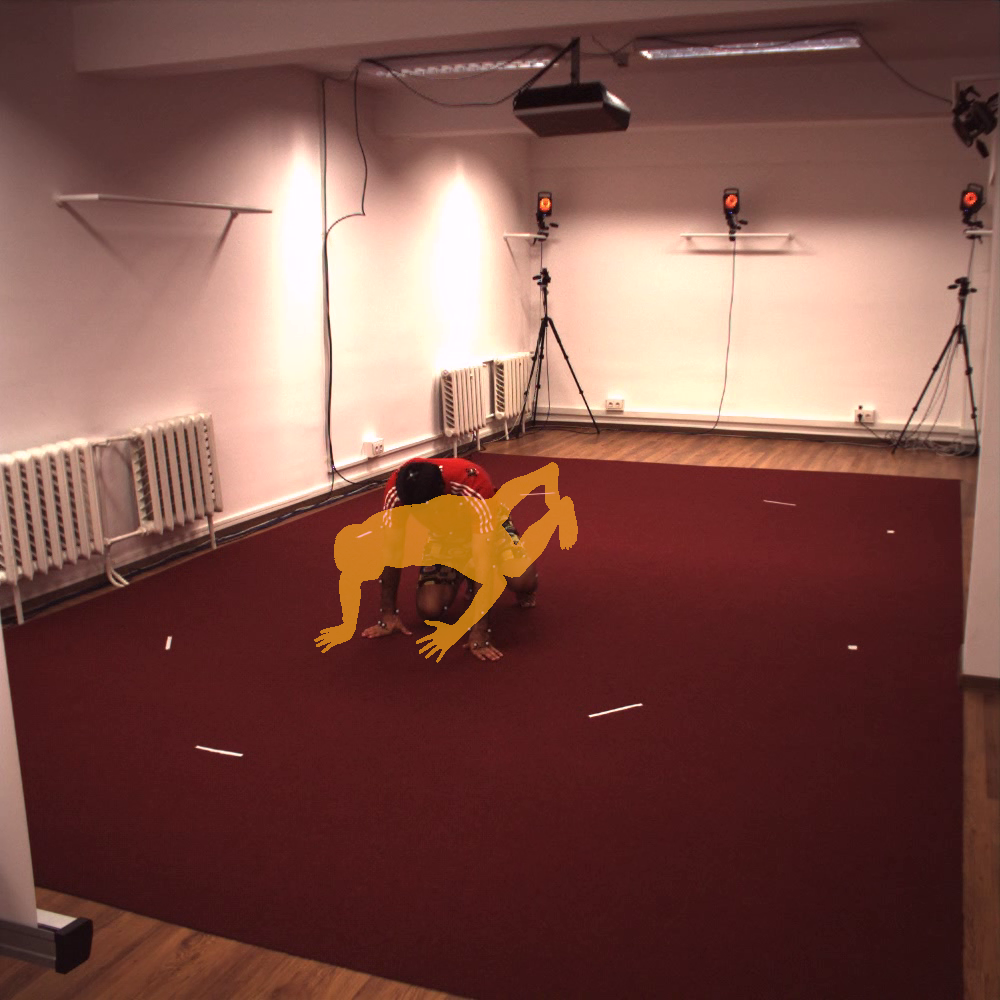}
\end{subfigure}\hfill
\begin{subfigure}{0.18\textwidth}
    \vspace{5pt}
    \includegraphics[width=\textwidth]{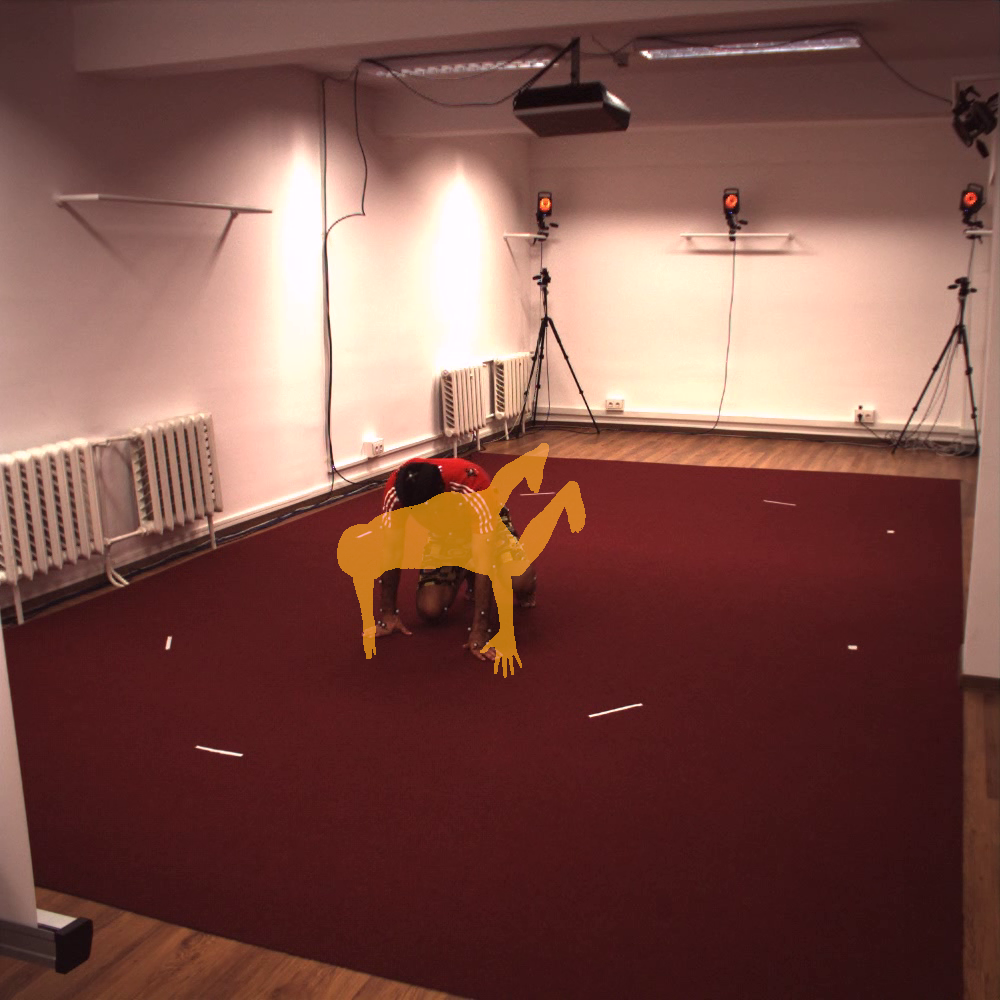}
\end{subfigure}\hfill
\begin{subfigure}{0.18\textwidth}
    \vspace{5pt}
    \includegraphics[width=\textwidth]{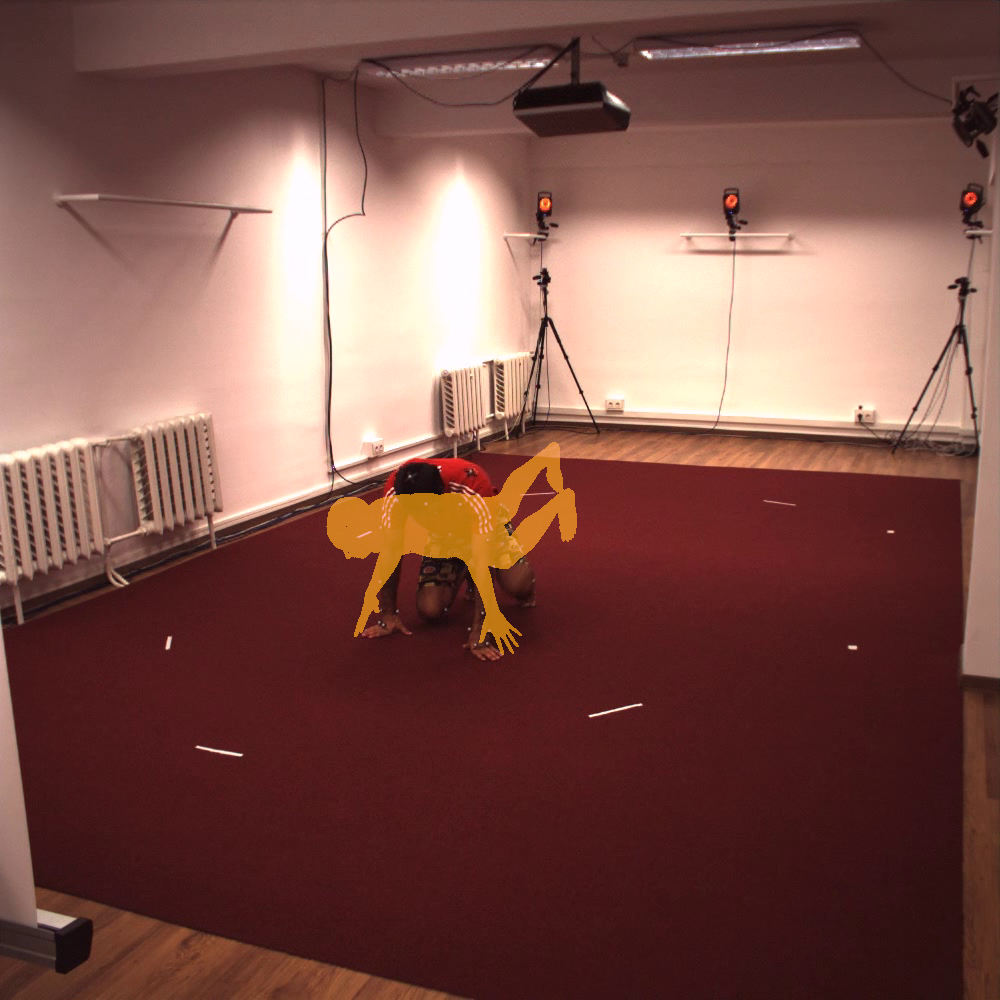}
\end{subfigure}\hfill
\begin{subfigure}{0.18\textwidth}
    \vspace{5pt}
    \includegraphics[width=\textwidth]{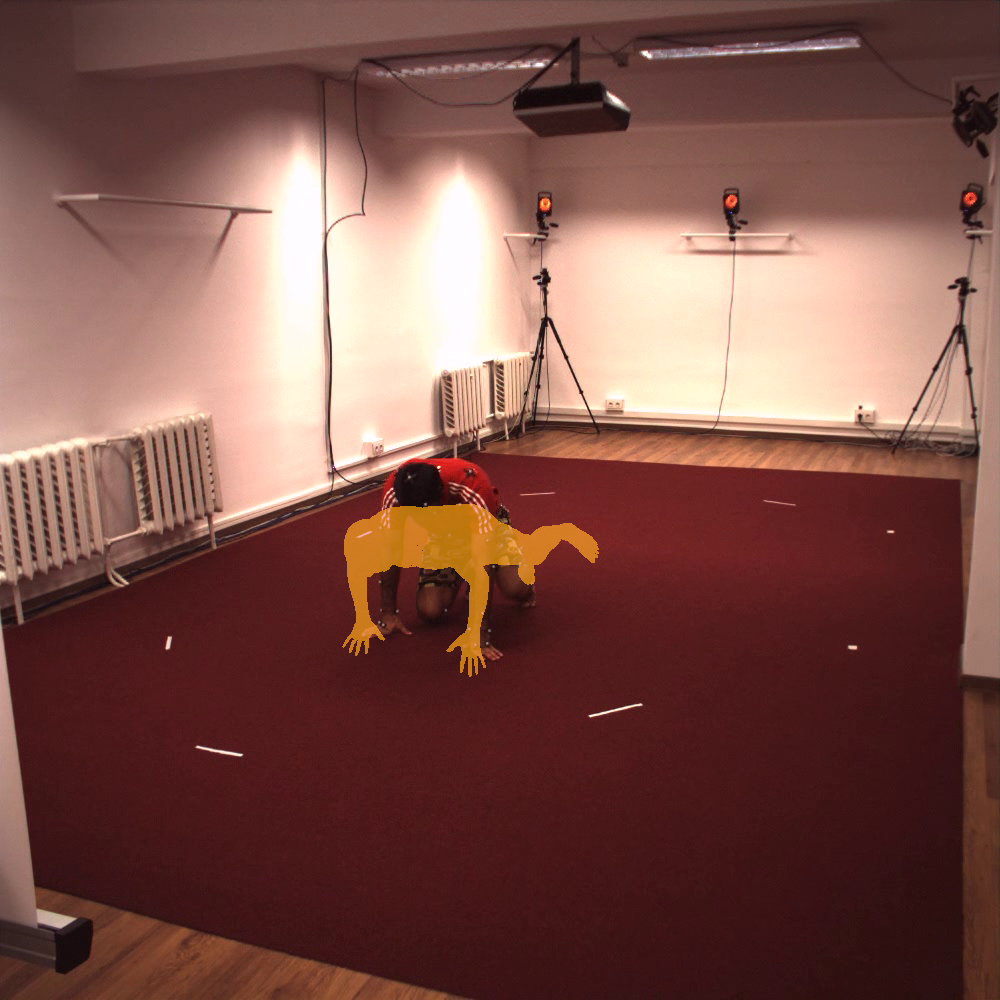}
\end{subfigure}

\begin{subfigure}{0.18\textwidth}
    \vspace{5pt}
    \includegraphics[width=\textwidth]{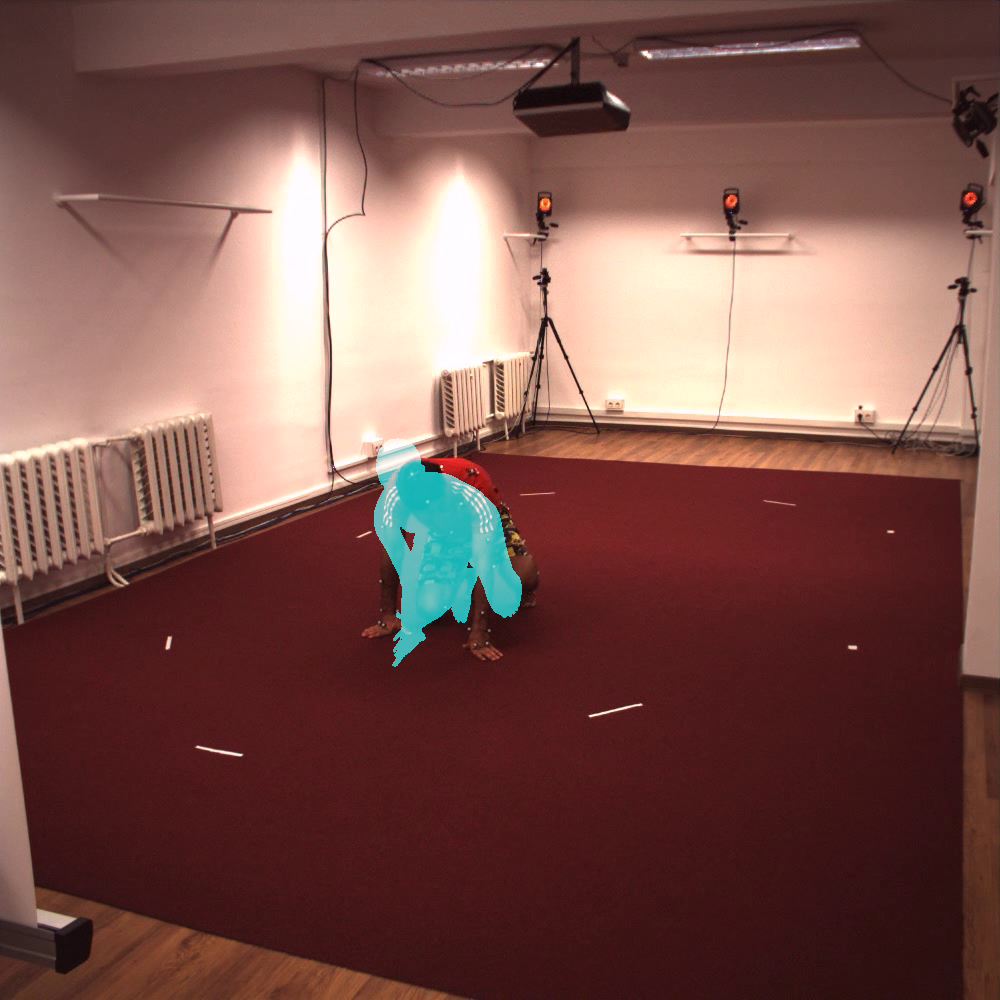}
\end{subfigure}\hfill
\begin{subfigure}{0.18\textwidth}
    \vspace{5pt}
    \includegraphics[width=\textwidth]{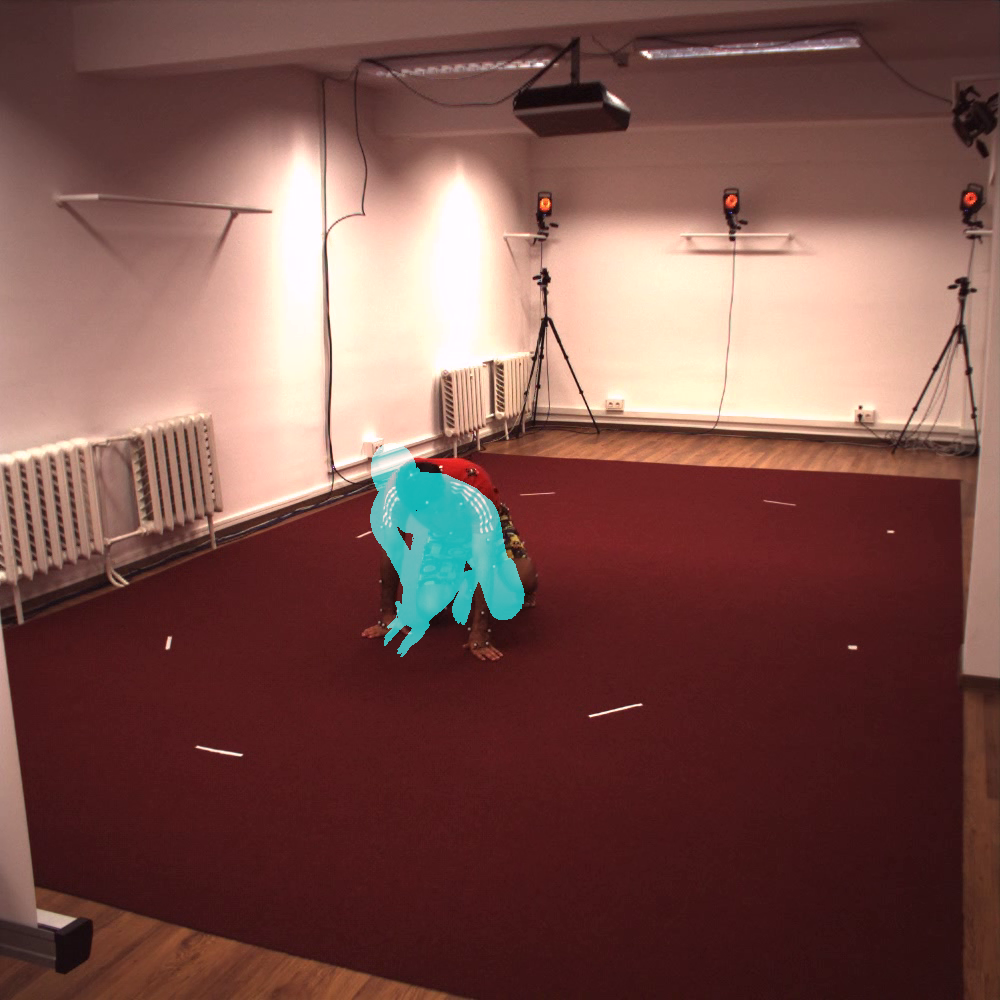}
\end{subfigure}\hfill
\begin{subfigure}{0.18\textwidth}
    \vspace{5pt}
    \includegraphics[width=\textwidth]{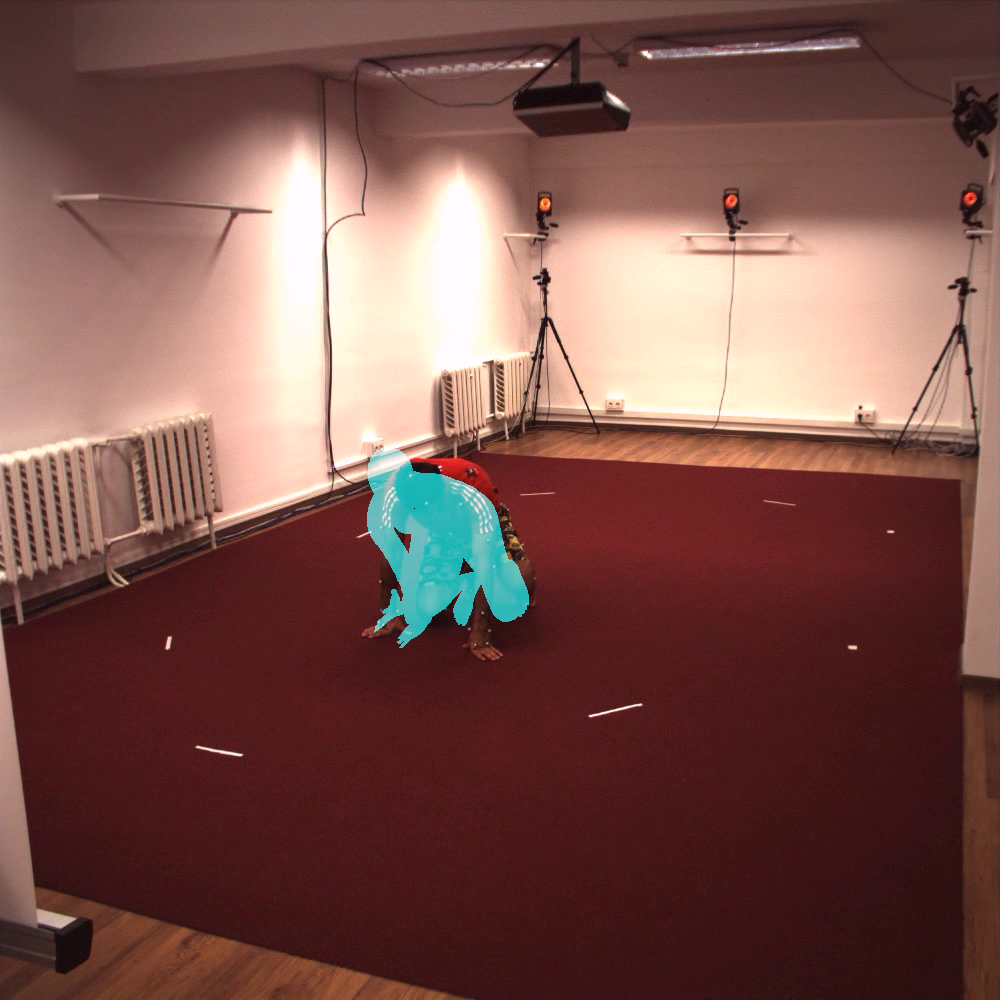}
\end{subfigure}\hfill
\begin{subfigure}{0.18\textwidth}
    \vspace{5pt}
    \includegraphics[width=\textwidth]{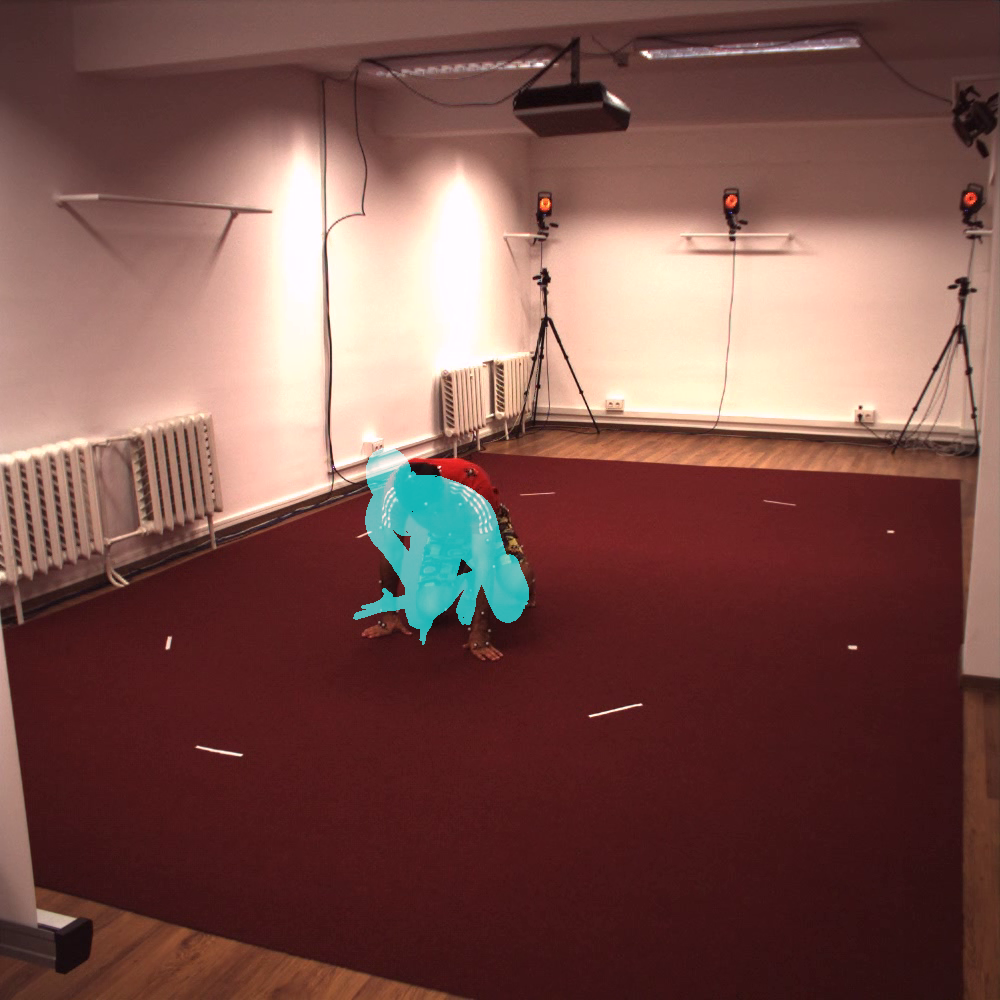}
\end{subfigure}\hfill
\begin{subfigure}{0.18\textwidth}
    \vspace{5pt}
    \includegraphics[width=\textwidth]{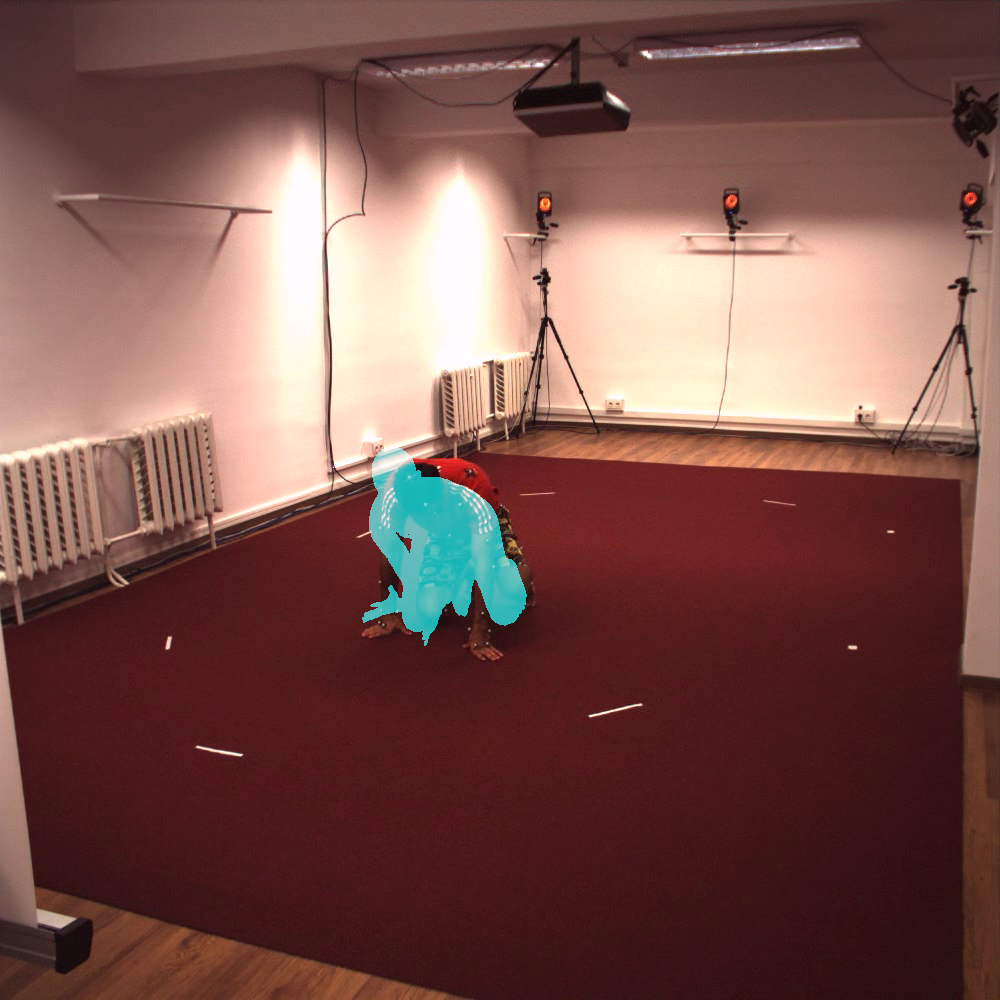}
\end{subfigure}

\begin{subfigure}{0.18\textwidth}
    \vspace{5pt}
    \includegraphics[width=\textwidth]{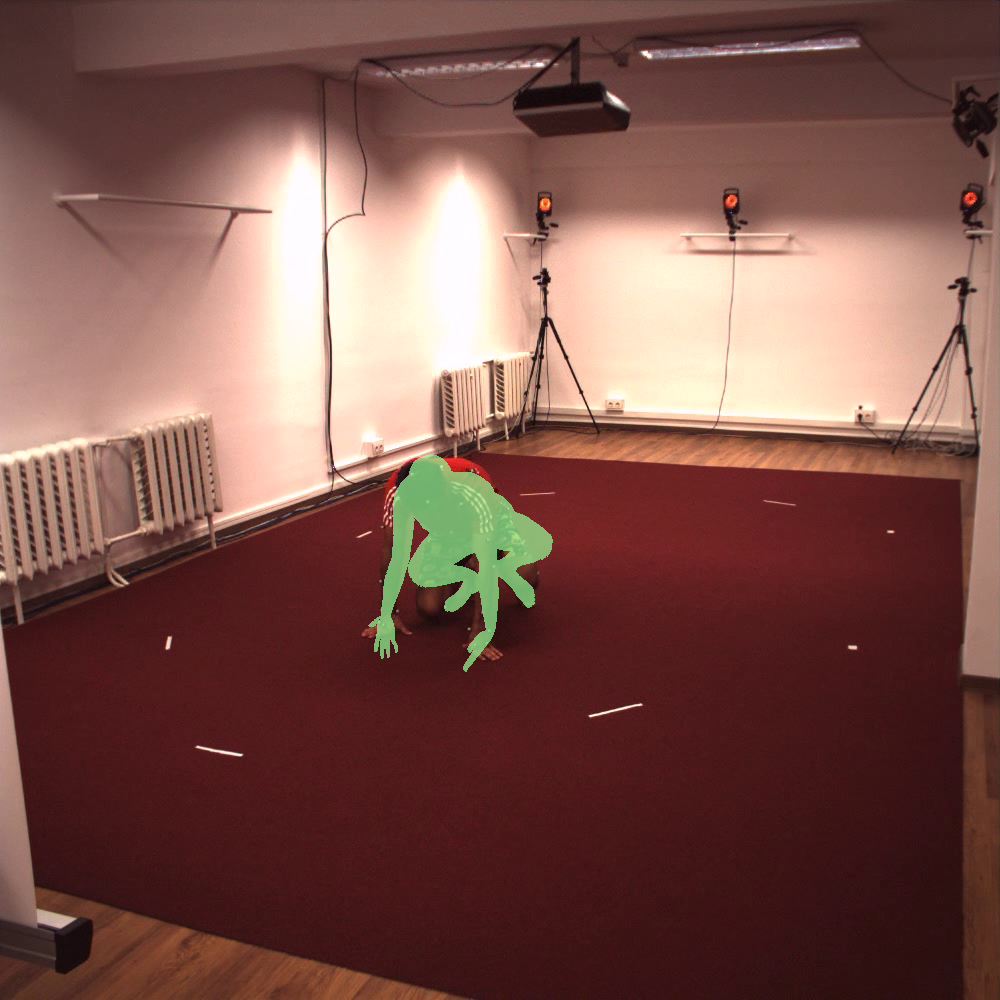}
\end{subfigure}\hfill
\begin{subfigure}{0.18\textwidth}
    \vspace{5pt}
    \includegraphics[width=\textwidth]{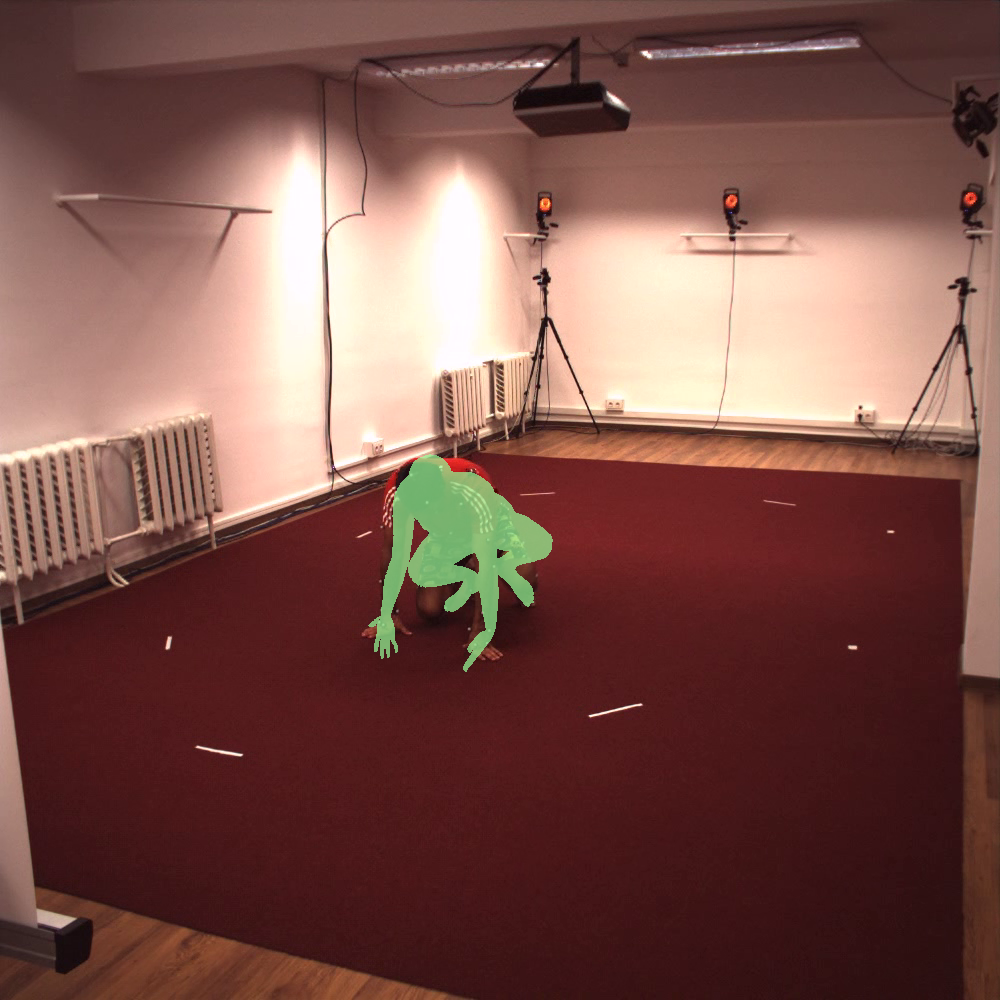}
\end{subfigure}\hfill
\begin{subfigure}{0.18\textwidth}
    \vspace{5pt}
    \includegraphics[width=\textwidth]{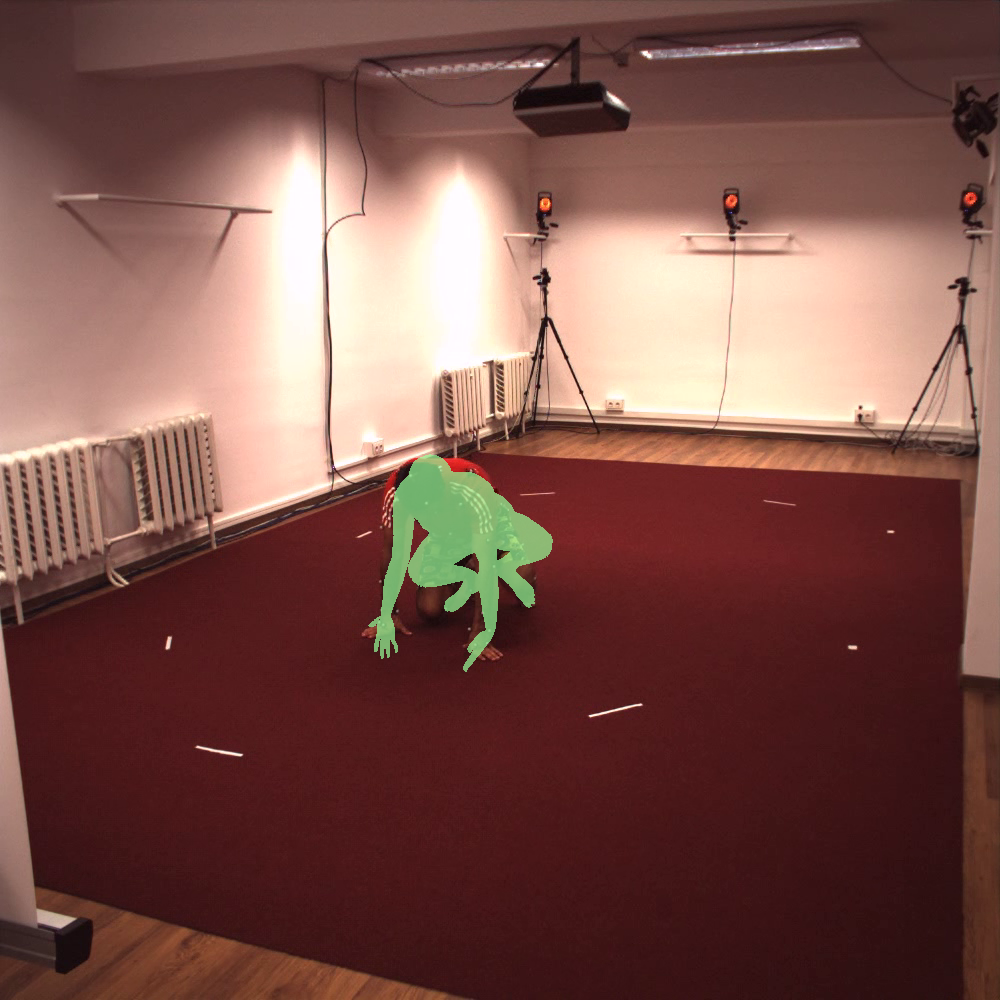}
\end{subfigure}\hfill
\begin{subfigure}{0.18\textwidth}
    \vspace{5pt}
    \includegraphics[width=\textwidth]{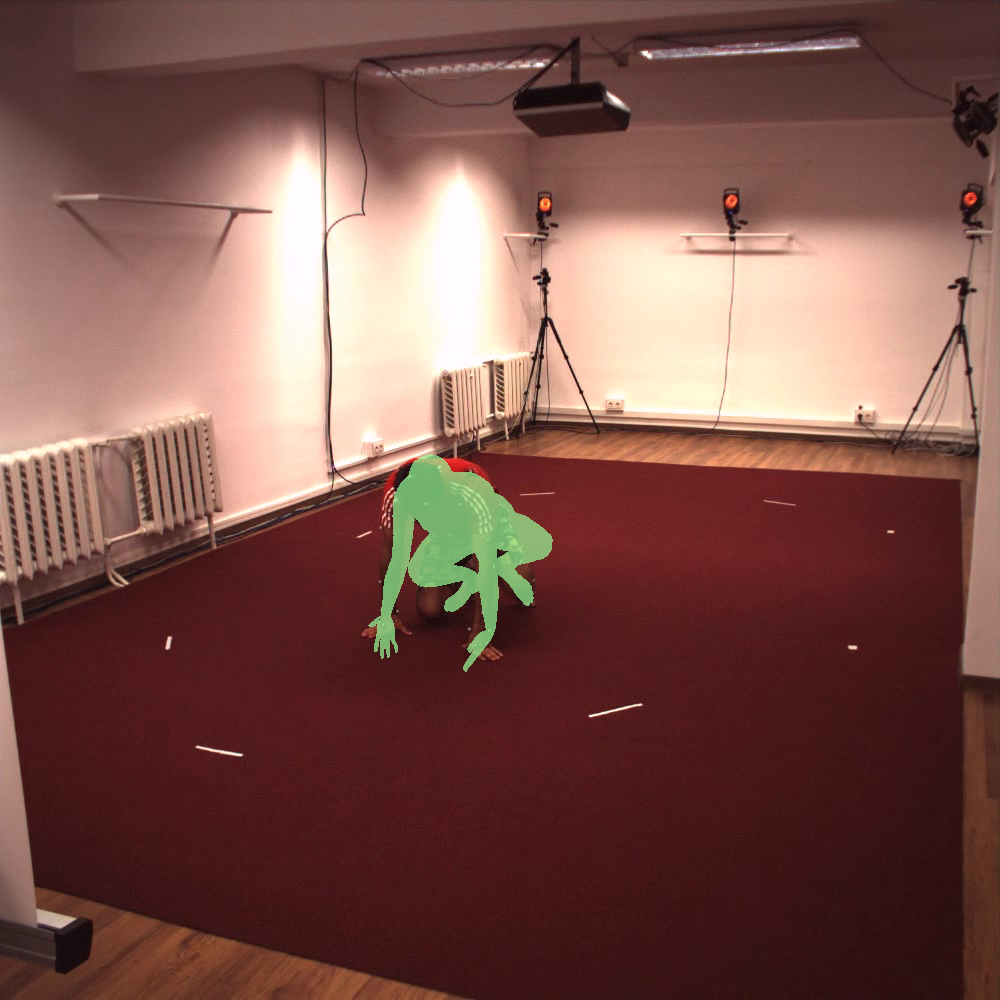}
\end{subfigure}\hfill
\begin{subfigure}{0.18\textwidth}
    \vspace{5pt}
    \includegraphics[width=\textwidth]{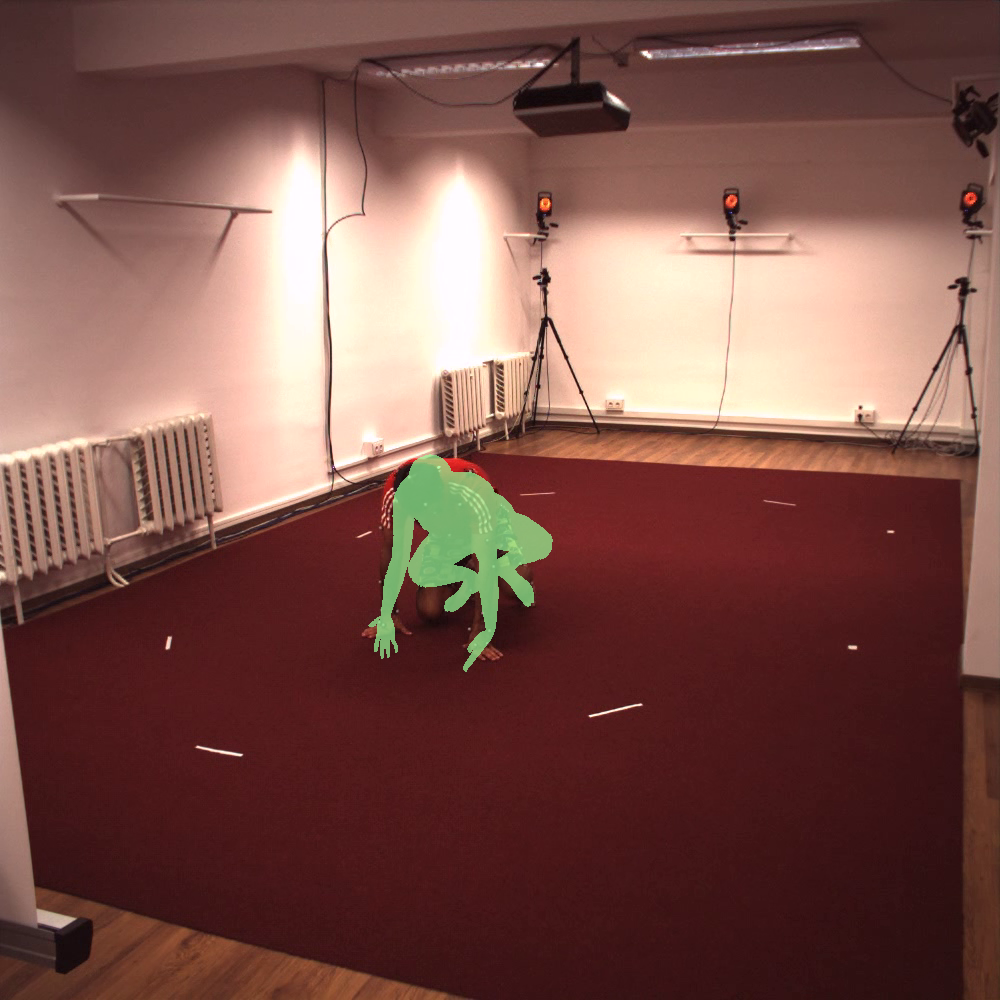}
\end{subfigure}

\begin{subfigure}{0.18\textwidth}
    \vspace{5pt}
    \includegraphics[width=\textwidth]{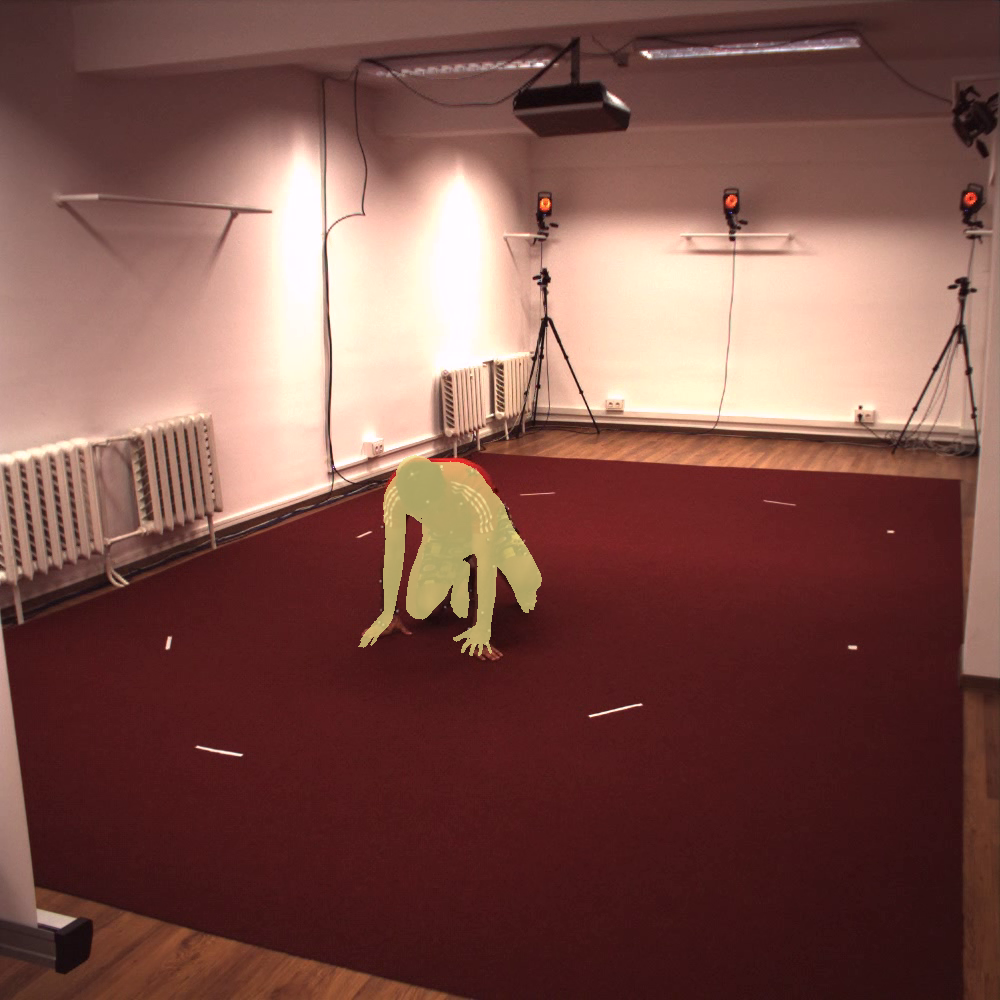}
\end{subfigure}\hfill
\begin{subfigure}{0.18\textwidth}
    \vspace{5pt}
    \includegraphics[width=\textwidth]{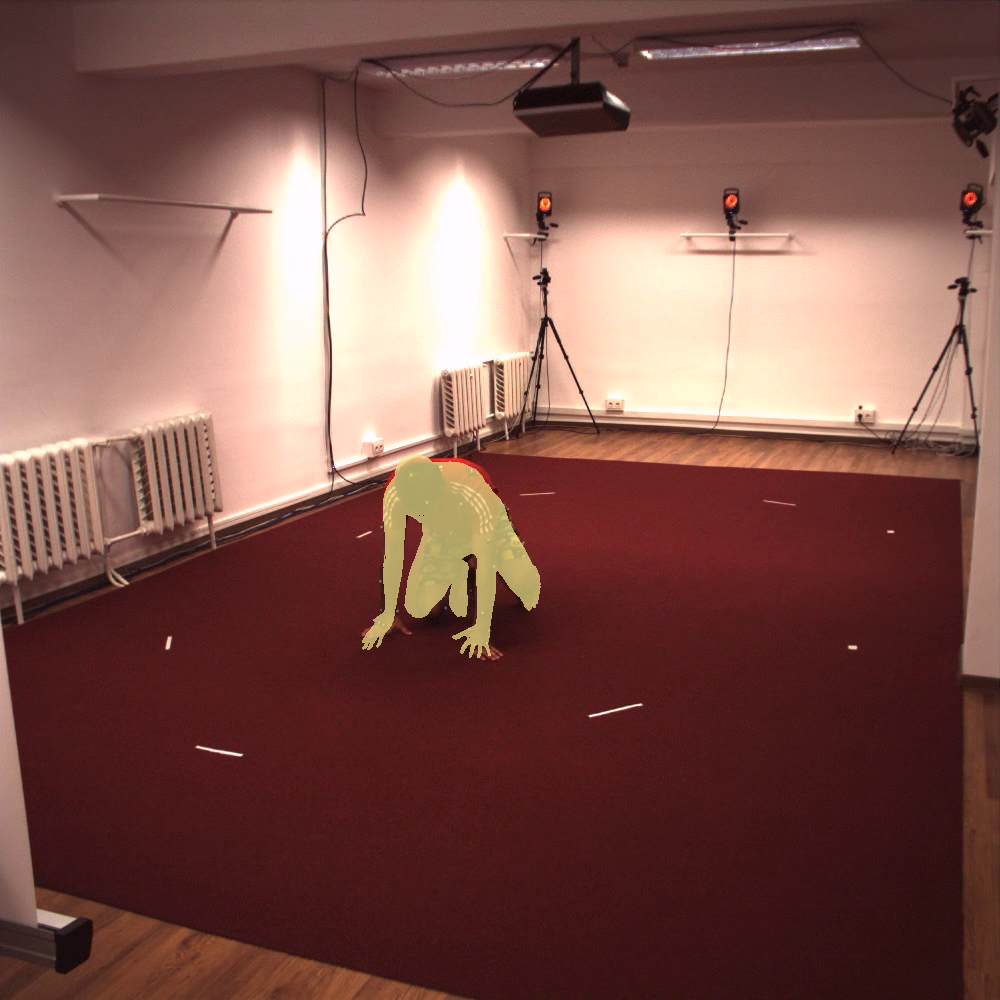}
\end{subfigure}\hfill
\begin{subfigure}{0.18\textwidth}
    \vspace{5pt}
    \includegraphics[width=\textwidth]{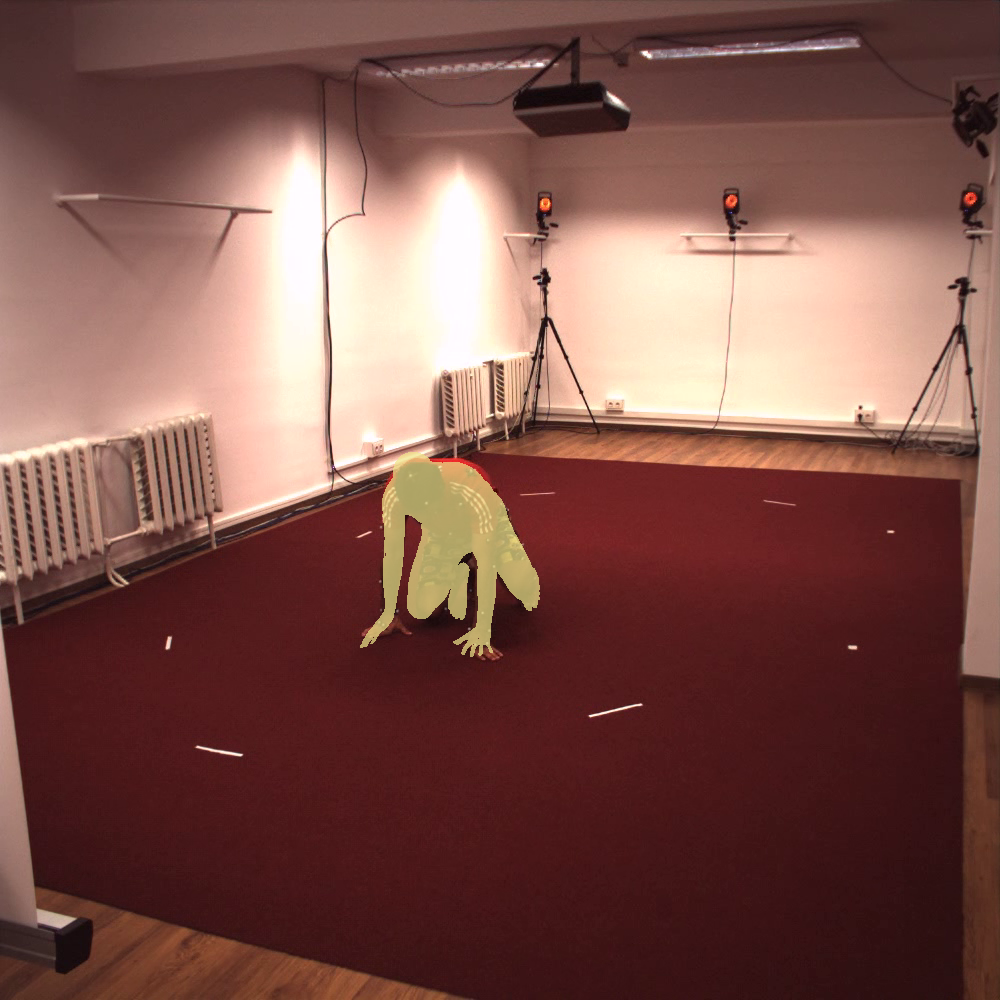}
\end{subfigure}\hfill
\begin{subfigure}{0.18\textwidth}
    \vspace{5pt}
    \includegraphics[width=\textwidth]{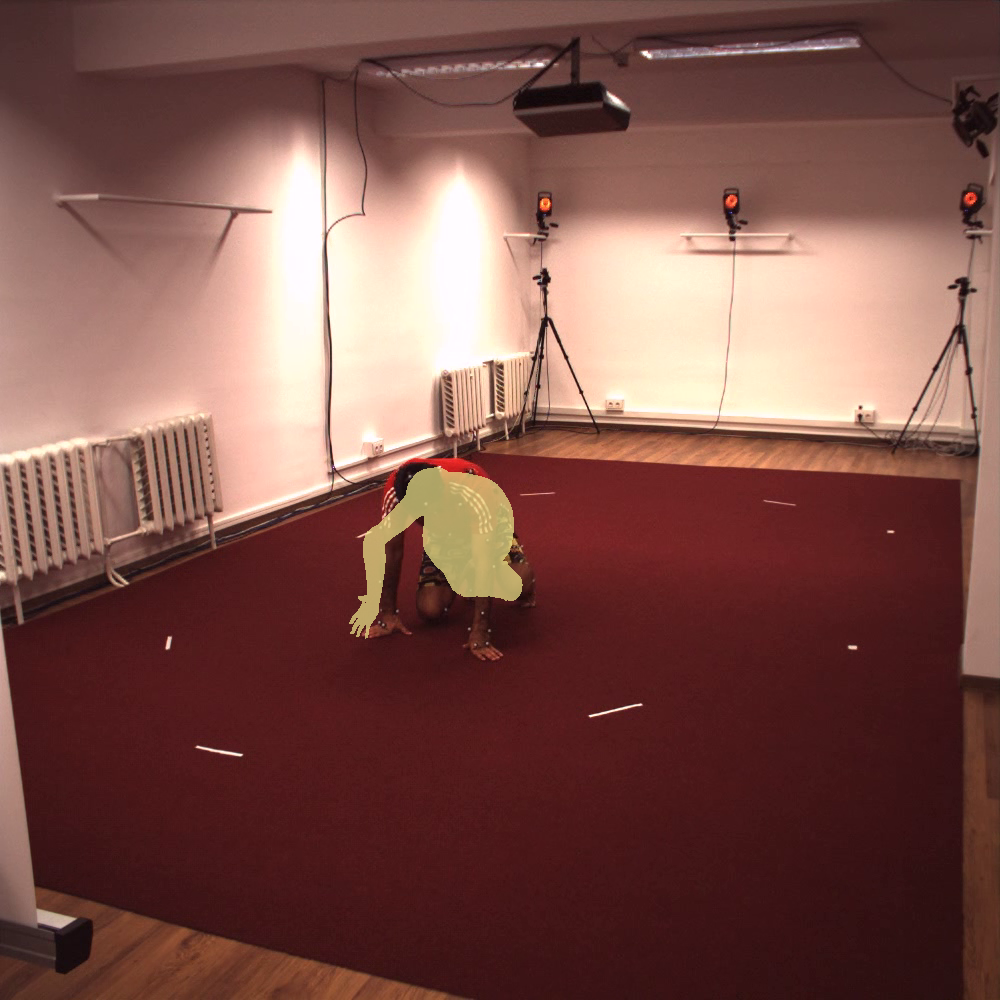}
\end{subfigure}\hfill
\begin{subfigure}{0.18\textwidth}
    \vspace{5pt}
    \includegraphics[width=\textwidth]{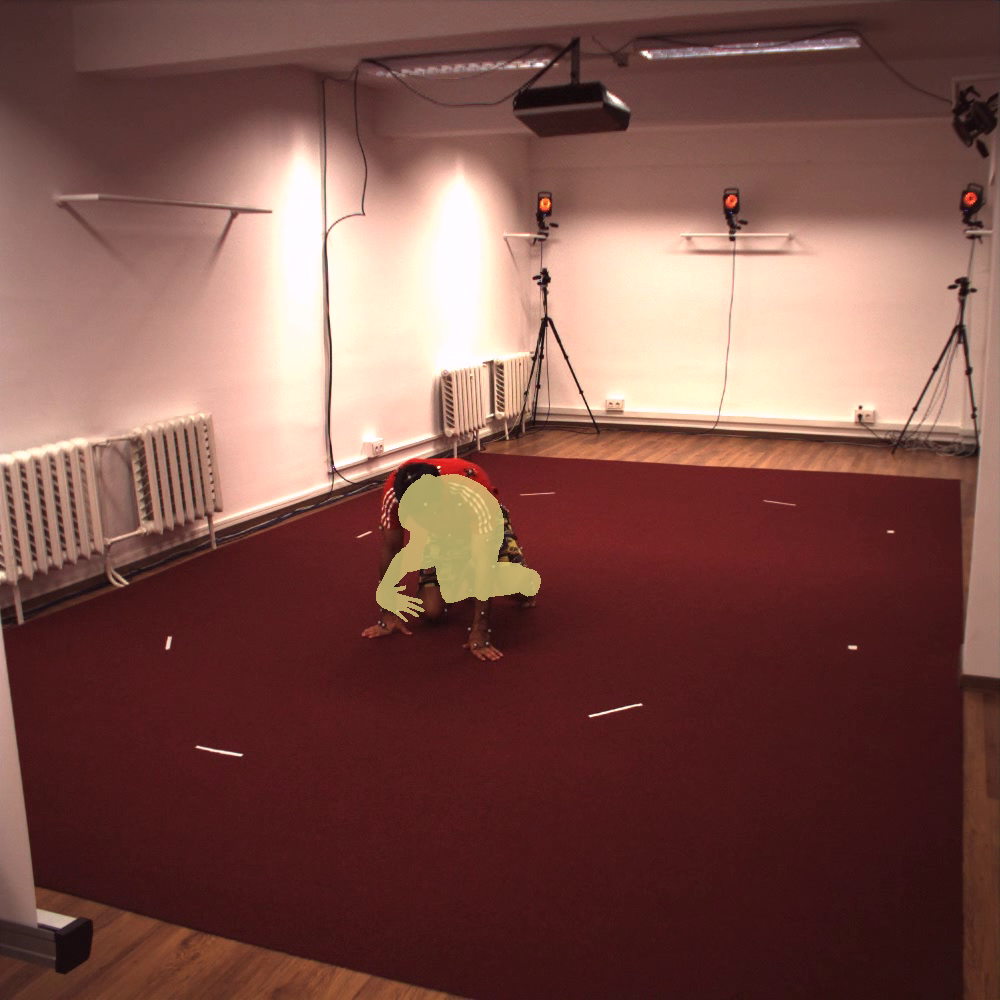}
\end{subfigure}

\begin{subfigure}{0.18\textwidth}
    \vspace{5pt}
    \includegraphics[width=\textwidth]{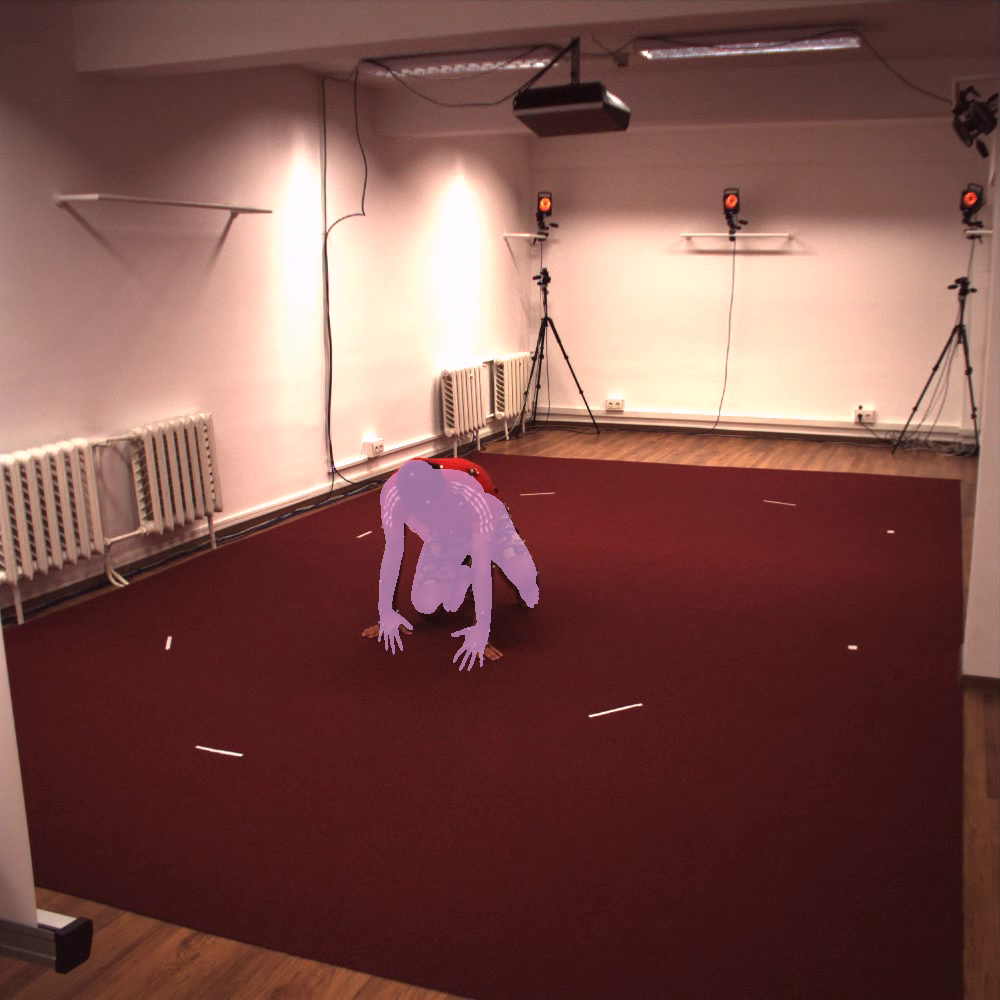}
\end{subfigure}\hfill
\begin{subfigure}{0.18\textwidth}
    \vspace{5pt}
    \includegraphics[width=\textwidth]{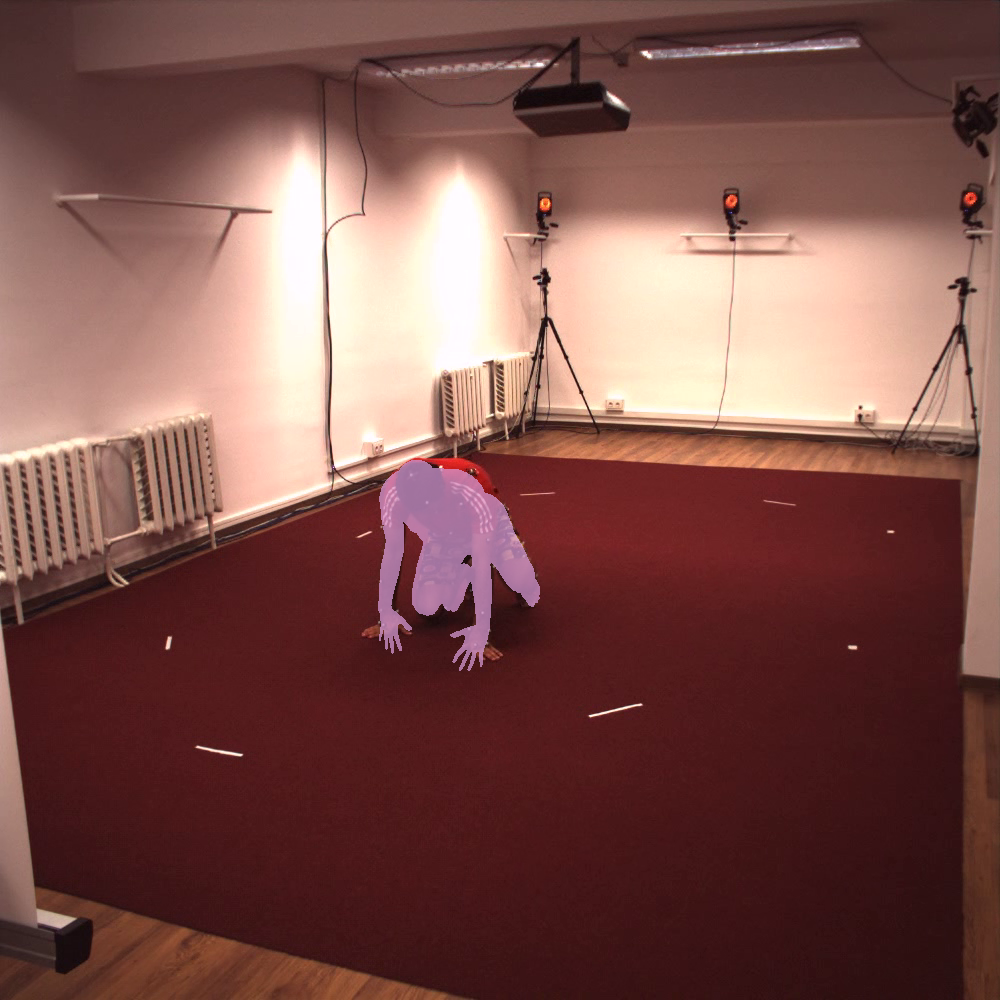}
\end{subfigure}\hfill
\begin{subfigure}{0.18\textwidth}
    \vspace{5pt}
    \includegraphics[width=\textwidth]{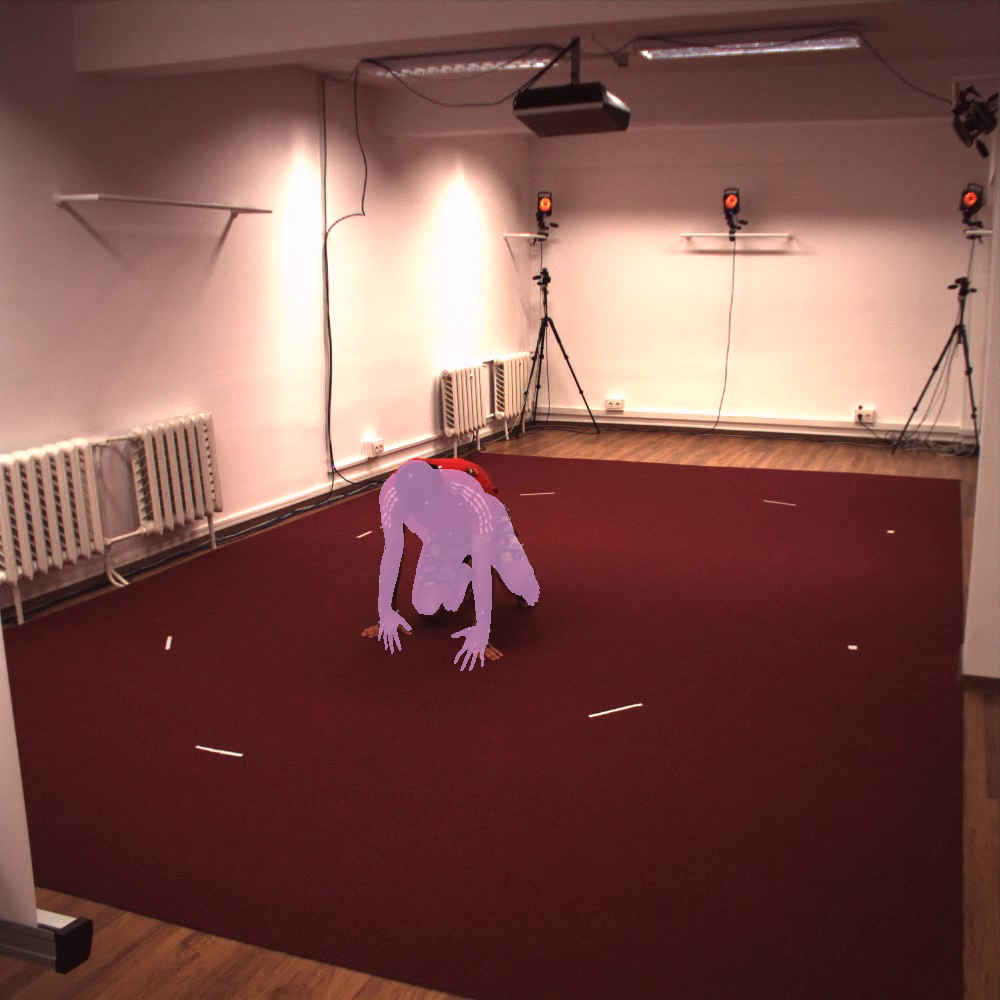}
\end{subfigure}\hfill
\begin{subfigure}{0.18\textwidth}
    \vspace{5pt}
    \includegraphics[width=\textwidth]{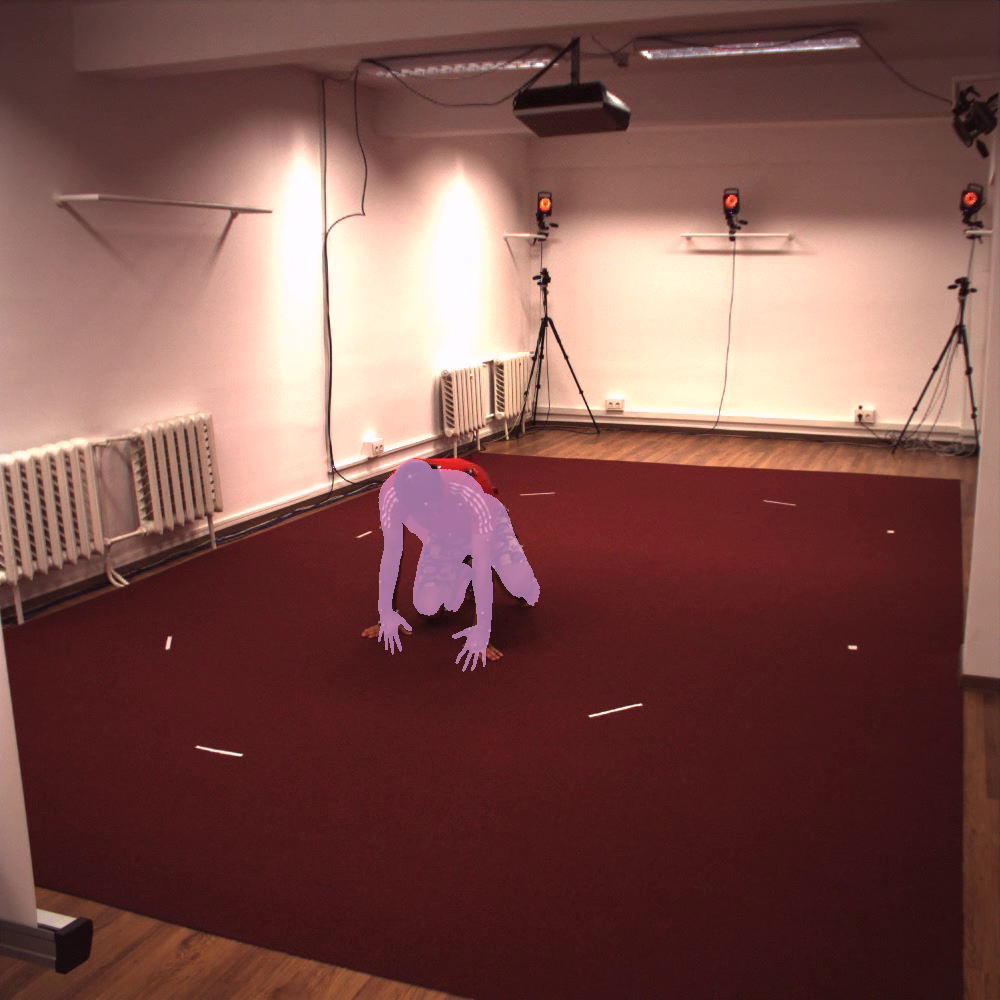}
\end{subfigure}\hfill
\begin{subfigure}{0.18\textwidth}
    \vspace{5pt}
    \includegraphics[width=\textwidth]{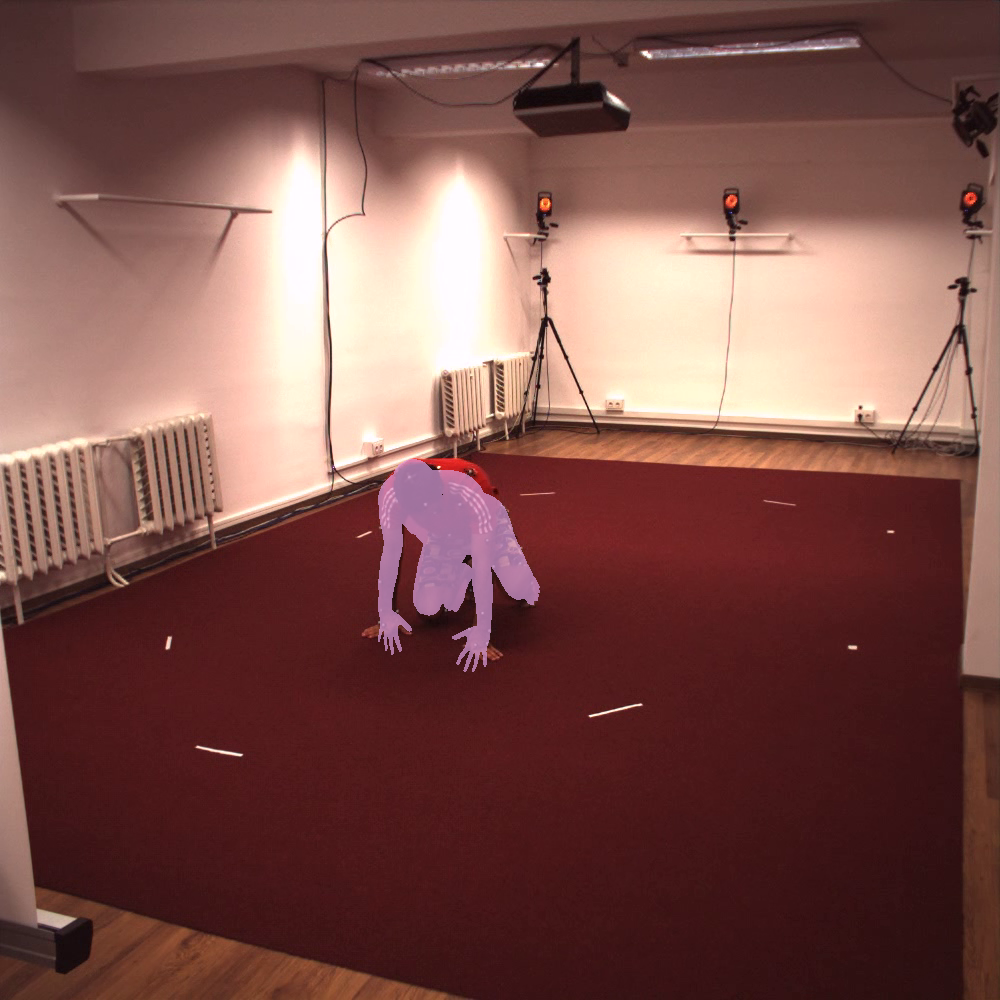}
\end{subfigure}
    
\caption{Qualitative comparison on Human3.6M for a sequence of five frames (columns, from left to right), of the SittingDown action of subject S9. 
The rows from top to bottom depict results from ${MuVS}$ (gray), 
DMMR \cite{huang2021dynamic} (\textcolor{orange}{orange}), SLAHMR \cite{ye2023decoupling} (\textcolor{cyan}{cyan}), DCT \cite{huang2017towards} (\textcolor{green}{green}), ETC \cite{arnab2019exploiting} (\textcolor{yellow}{yellow}), and last BundleMoCap (Ours - \textcolor{magenta}{magenta}) . 
Our approach is robust to occlusions and wrong keypoints estimations that greatly influence other approaches, especially under sparse views.
This allows BundleMoCap to accurately capture human motion despite potential obstacles and inaccuracies in the estimated data, demonstrating its effectiveness and reliability in complex scenarios.
}
\label{fig:results}

\end{figure*}

\subsubsection{\textit{Performance and robustness to outliers}}
Tables~\ref{tab:results_h36m} and \ref{tab:results_3dhp} present the quantitative results on the Human3.6M and MPI-INF-3DHP datasets respectively.
Our BundleMoCap approach outperforms all other methods that optimize and solve over multiple stages and use motion smoothness constraints. 
Notably, BundleMoCap relies on a pose manifold and its local smoothness to reconstruct local smooth motion via manifold interpolation, compared to the autoregressive HuMoR temporal prior \cite{rempe2021humor,ye2023decoupling} (SLAHMR) or the recurrent VPoser-t motion prior \cite{huang2021dynamic} that reconstruct motion segments (DMMR).
Further, all other bundle solving methods \cite{huang2017towards,huang2021dynamic,arnab2019exploiting,ye2023decoupling} solve multiple frames in a bundle, whereas BundleMoCap solves for a single latent keyframe, while reconstructing the intermediate frames.
Crucially, this property offers higher robustness to outliers, as it is not possible to interpolate to a spurious pose that would manifest due to erroneous/conflicting multi-view keypoint estimates. 
Such cases are illustrated in Figure~\ref{fig:results} in a sequence that contains segments suffering from outlier keypoint estimates.
These result in temporally inconsistent results for all other methods, compared to the BundleMoCap results that remain temporally coherent.

\begin{table}[!htbp]
\centering
\caption{Quantitative comparison against other methods on the Human3.6M dataset. 
The average error/accuracy over all actions is reported.
\first{Bold red} marks the best performing row, \second{orange} the second best, and \third{yellow} the third.
The arrows next to each metric denote the direction of better performance.
}
\label{tab:results_h36m}
\resizebox{\columnwidth}{!}{%
\begin{tabular}{c|ccccc}
\hline
                               & \down{MPJPE}        & \down{RMSE}         & \down{MAE}              & \up{PCK3}        & \up{PCK7}        \\ \hline
${MuVS}$                       & 43.83  $mm$         & 48.33  $mm$         & 4.56$^{\circ}$          & 40.58\%          & 93.75\%          \\
DCT \cite{huang2017towards}    & 42.34 $mm$          & 48.13 $mm$          & 4.18$^{\circ}$          & \third{40.84\%}          & \second{94.97\%} \\
DMMR \cite{huang2021dynamic}   & 41.3 $mm$           & 47.66 $mm$          & 4.76$^{\circ}$          & 39.52\%          & 92.11\%          \\
SLAHMR \cite{ye2023decoupling} & \third{40.8} $mm$   & \third{42.66} $mm$  & \third{4.06}$^{\circ}$  & \second{40.86\%} & \second{94.97\%} \\
ETC \cite{arnab2019exploiting} & \second{37.51} $mm$ & \second{41.53} $mm$ & \second{4.05}$^{\circ}$ & 40.30\%          & \third{94.50\%}  \\
BundleMoCap (Ours)             & \first{36.48} $mm$  & \first{40.34} $mm$  & \first{3.96}$^{\circ}$  & \first{41.16\%}  & \first{95.71\%}  \\ \hline
\end{tabular}
}
\normalsize
\end{table}
\begin{table}[!htbp]
\centering
\caption{Quantitative comparison against other methods on the MPI-INF-3DHP dataset. 
The average error/accuracy over all actions is reported.
\first{Bold red} marks the best performing row, \second{orange} the second best, and \third{yellow} the third.
The arrows next to each metric denote the direction of better performance.
}
\label{tab:results_3dhp}
\resizebox{\columnwidth}{!}{%
\begin{tabular}{c|ccccc}
\hline
                             & \down{MPJPE}        & \down{RMSE} & \down{MAE}     & \up{PCK3} & \up{PCK7} \\ \hline
${MuVS}$                     & 64.99 $mm$          & 76.12  $mm$ & 6.28$^{\circ}$ & 28.20\%   & 73.75\%   \\
DCT \cite{huang2017towards}  & 62.43 $mm$          & 68.13 $mm$  & 6.18$^{\circ}$ & 35.84\%   & 83.77\%   \\
DMMR \cite{huang2021dynamic} & \second{57.51} $mm$ & 67.66 $mm$  & 6.06$^{\circ}$ & 37.52\%   & 81.11\%   \\
SLAHMR \cite{ye2023decoupling} & 61.8 $mm$          & \third{62.55} $mm$  & \third{5.74}$^{\circ}$  & \second{40.86\%} & \third{83.97\%}  \\
ETC \cite{arnab2019exploiting} & \third{59.51} $mm$ & \second{61.32} $mm$ & \second{5.64}$^{\circ}$ & \third{39.30\%}  & \second{84.50\%} \\
BundleMoCap (Ours)             & \first{56.41} $mm$ & \first{59.12} $mm$  & \first{5.43}$^{\circ}$  & \first{44.51\%}  & \first{85.71\%}  \\ \hline
\end{tabular}
}
\normalsize
\end{table}
\begin{figure}[!htbp]
\includegraphics[width=\linewidth]{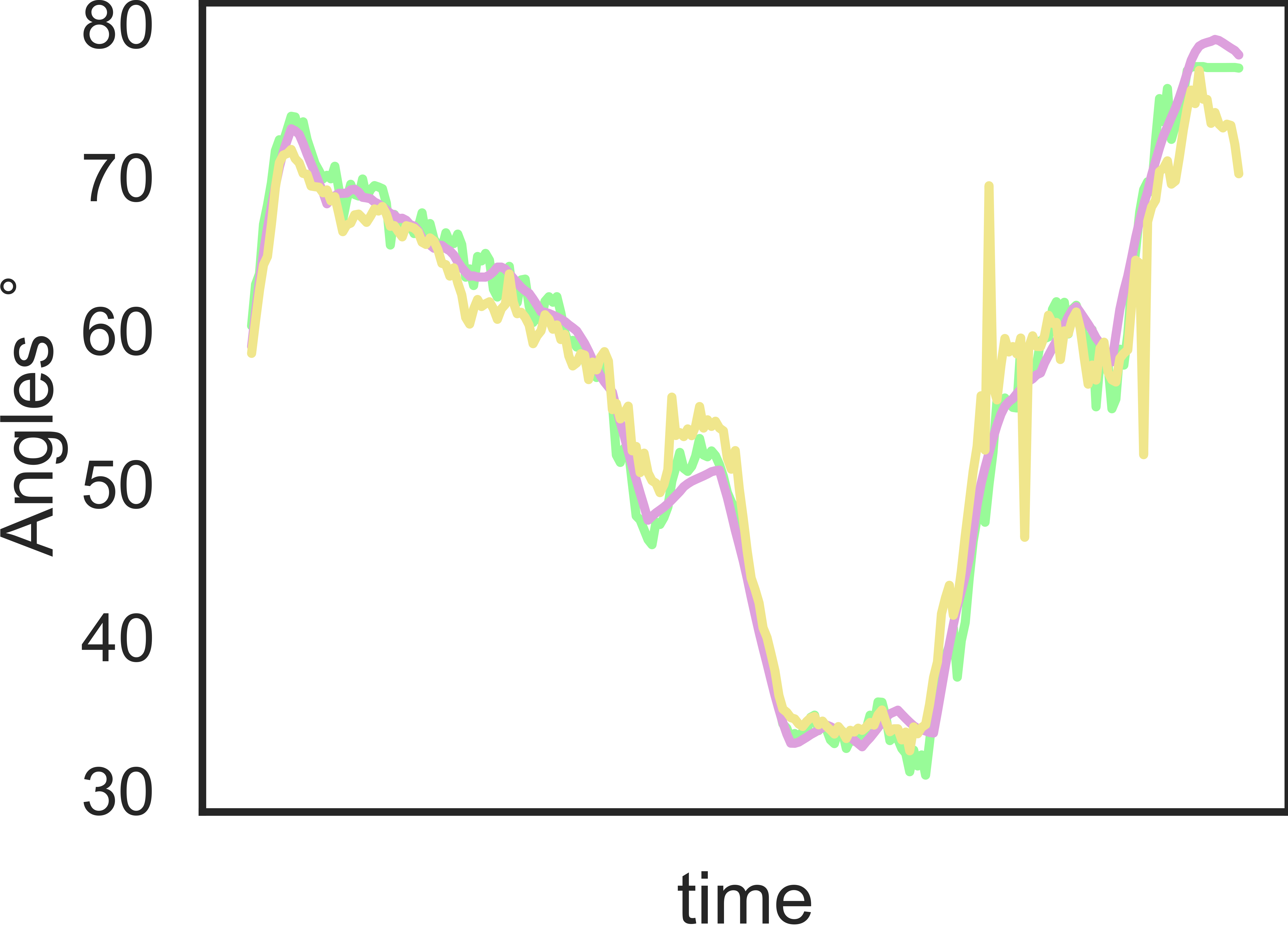}
\caption{
A segment of the knee flexion angle for the \textit{Sitting Down} action performed by S9 in the Human3.6M dataset.
Despite not enforcing temporal consistency during optimization, BundleMoCap delivers smooth motions comparable to other state-of-the-art techniques (DCT \cite{huang2017towards} - \textcolor{green}{green}, ETC \cite{arnab2019exploiting} - \textcolor{yellow}{yellow}) that include smooth motion objectives in their optimization formulations. 
This can be attributed to the expressiveness of the pose manifold $\manifold$, which facilitates locally smooth transitions across poses, enabling our method to capture motions fluidly. 
Interestingly, this does not negatively impact performance, as evident from the results in Tables~\ref{tab:results_h36m} and \ref{tab:results_3dhp}.
}
\label{fig:smoothness}
\end{figure}

\subsubsection{\textit{Ensuring smooth motions}}
Apart from the temporal coherence, BundleMoCap delivers smooth motion captures, minimizing jitter, despite noisy keypoint estimates.
This is achieved without using any temporal smoothness objective, compared to the other methods using joint smoothness \cite{huang2017towards,arnab2019exploiting,ye2023decoupling}, or latent code smoothness \cite{huang2021dynamic}.
To illustrate this point, we extract the knee flexion, an angle between three joints, namely the hip, knee and ankle.
Any jitter in these joints estimates will result in noisy angle extraction.
As evident in Figure~\ref{fig:smoothness} the angle extracted from the results of our method exhibits smoother motion compared to both a segment solver \cite{huang2017towards} as well as an entire sequence solver \cite{arnab2019exploiting}.
Additional results showcasing the smooth MoCap that BundleMoCap offers can be found in the supplementary video, either in standalone qualitative results or compared to the other approaches (using the same color coding as Figure~\ref{fig:results}).

\subsubsection{\textit{Runtime efficiency}}
\begin{figure}[!htbp]
\includegraphics[width=\linewidth]{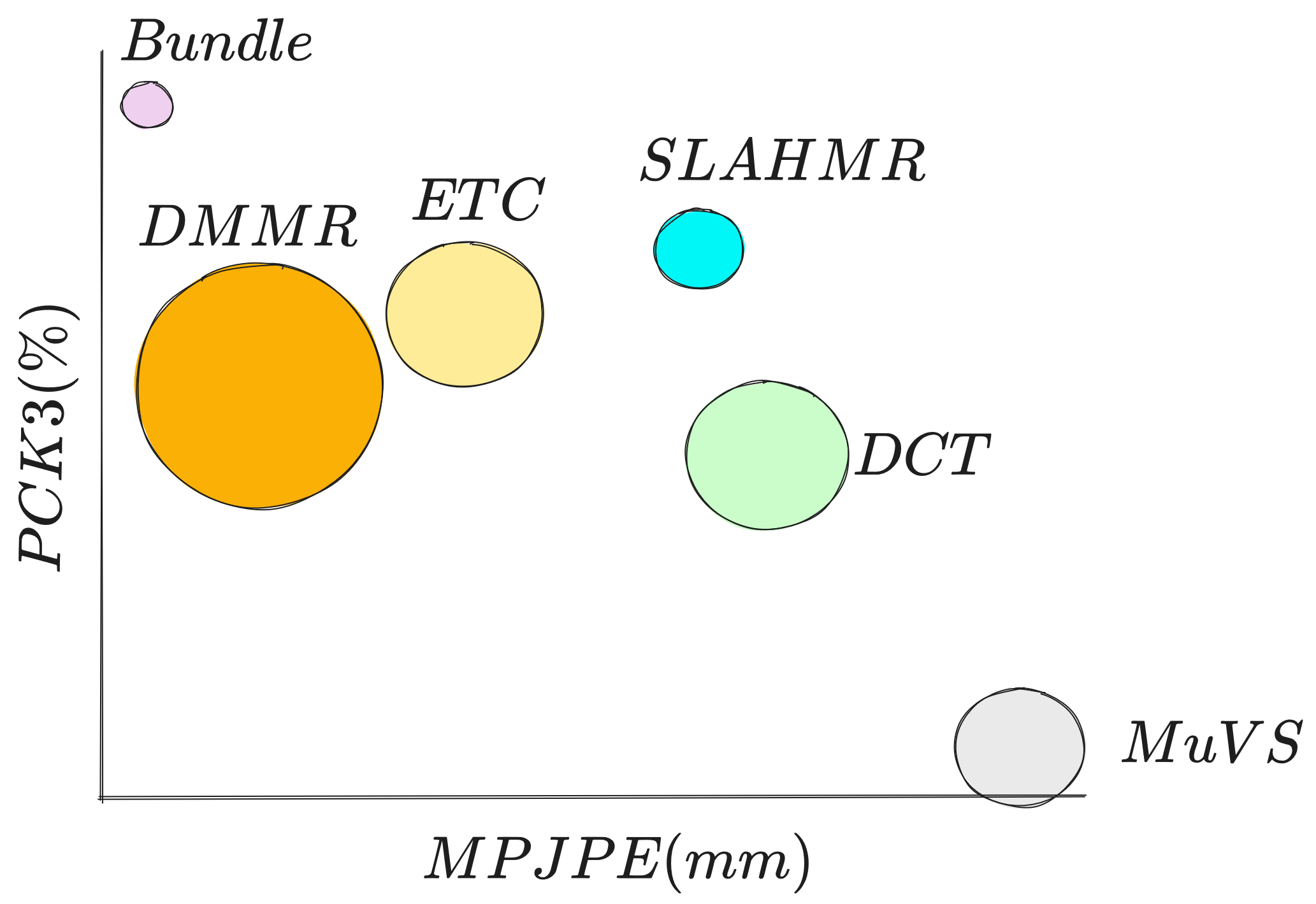}
\caption{
A visual summary of the performance (horizontal and vertices axes) and efficiency (runtime illustrated as the size of each point) of different methods. 
BundleMoCap's competitive results are achieved with a minimal computational burden, without the need for 3D initialization, or a smoothness objective. 
Its efficiency is greatly boosted from its single-stage nature, making it an more appropriate choice for practical applications.
}
\label{fig:efficiency}
\end{figure}

Finally, BundleMoCap is a highly efficient method for solving sparse multi-view sequences compared to other approaches whose complexity does not scale efficiently with the sequence size.
BundleMoCap solves a single/first frame in isolation and then solves for frame bundles of length $10$ using a single latent code.
In contrast, other approaches either require per-frame initialization before solving the entire sequence \cite{arnab2019exploiting}, single-frame fits over the entire sequence before solving bundles of frames \cite{huang2017towards,ye2023decoupling}, or solve bundles of frames whilst optimizing each frame's latent code across multiple ($4$) stages \cite{huang2021dynamic}.
BundleMoCap solves the entire sequence in a single-shot, using a single sliding window optimization stage.
Indicatively, the average optimization runtimes for each method can be outlined as follows:
BundleMoCap takes around 1 hour for processing a 4-view sequence of 2000 frames, while DMMR involves a longer process with 4 stages spanning almost 1 day (24 hours). 
ETC requires an initialization pass over the entire sequence, with the optimization process taking approximately 3 hours. 
For DCT, the initialization from ${MuVS}$ consumes around 40 minutes, and the subsequent optimization requires an additional 9 hours. 
Lastly, SLAHMR itself takes an average of 6 hours for optimization. 
It is essential to consider that these times are approximate and may differ based on the specific hardware and software configurations employed during the experiments. 
However, in our implementation the same software version and hardware were used for consistency.
Even though these timings do not take into account the keypoint detector and surely all implementations can be improved, the above analysis aims to showcase the complexity differences between methods requiring multiple passes over the videos, either for initialization or as multiple optimization stages, in addition to the gains of solving less or more parameters/frames.
Figure~\ref{fig:efficiency} provides a comprehensive overview of the performance of each method, alongside its efficiency which is illustrated by each point's size that represents its relative runtime. 

\subsubsection{\textit{Error Accumulation}}
Using a sliding window approach comes with the risk of errors accumulating from one temporal window to the next.
BundleMoCap is robust to such drifting due to its bundle solving nature.
Solving using constraints for the single keyframe only and then reconstructing the temporal window would suffer, when outliers manifest at that specific keyframe.
Instead, BundleMoCap is constrained by the entire temporal window, which reduces the chances of errors accumulating across the entire sequence.
This is experimentally verified in Table~\ref{tab:results_3dhp} where the sequences solved span minutes.
In such long sequences drifting would hurt performance, but instead, BundleMoCap achieves good quantitative results.
Still, BundleMoCap crucially relies on the assumption that small paths on the manifold can reconstruct small motions.
It remains to be investigated if the expressivity of the manifold is hindered when bundle solving larger temporal windows and how that relates to the accumulation of errors.

\section{Conclusion}
In this work we have presented a novel method for solving MoCap in sparse multi-view videos.
Exploiting manifold interpolation, we solve for a bundle of frames reconstructed via two latent keyframes.
The resulting method is efficient, produces smooth motions and exhibits robustness to outlier observations.
However, it assumes that linear interpolation steps on the manifold correspond to linear pose space displacements.
While this may hold for some temporal window lengths, it remains to be investigated if this assumption holds for smaller/larger ones.
Similarly, BundleMoCap relies on high quality local manifold transitions, a trait that can be improved using different generative models. Still, the diversity of generations models' learned distributions may be limited by their training data.
Closing, while spherical interpolation proved to be sufficient, more involved interpolation schemes may allow for extending the temporal window, or boost solving accuracy.
Overall, we believe that there is opportunity for further investigation in this line of bundle MoCap solving.

\bibliographystyle{ACM-Reference-Format}
\bibliography{egbib}

\end{document}